%% file: main.tex
\documentclass[fleqn,10pt]{wlscirep}
\usepackage[utf8]{inputenc}
\usepackage[T1]{fontenc}
\usepackage{multirow} 
\usepackage{array}
\usepackage{graphicx}
\usepackage{subcaption}
\usepackage[acronym]{glossaries}
\usepackage{float}
\usepackage{placeins}
\FloatBarrier

\title{AI-assisted radiographic analysis in detecting alveolar bone-loss severity and patterns}

\author[1, +]{Chathura Wimalasiri}
\author[1, +]{Piumal Rathnayake}
\author[1, +]{Shamod Wijerathne}
\author[2]{Sumudu Rasnayaka}
\author[2]{Dhanushka Leuke Bandara}
\author[1]{Roshan G. Ragel}
\author[3]{Vajira Thambawita}
\author[1]{Isuru Nawinne}

\input{acronyms}

\affil[1]{Faculty of Engineering, University of Peradeniya, 20400, Sri Lanka}
\affil[2]{Faculty of Dental Sciences, University of Peradeniya, 20400, Sri Lanka}
\affil[3]{Simula Metropolitan Center for Digital Engineering, 0164 Oslo, Norway}


\begin{abstract}

Periodontitis, a chronic inflammatory disease causing alveolar bone loss, significantly affects oral health and quality of life. Accurate assessment of bone loss severity and pattern is critical for diagnosis and treatment planning. In this study, we propose a novel AI-based deep learning framework to automatically detect and quantify alveolar bone loss and its patterns using intraoral periapical (IOPA) radiographs. Our method combines YOLOv8 for tooth detection with Keypoint R-CNN models to identify anatomical landmarks, enabling precise calculation of bone loss severity. Additionally, YOLOv8x-seg models segment bone levels and tooth masks to determine bone loss patterns (horizontal vs. angular) via geometric analysis. Evaluated on a large, expertly annotated dataset of $1000$ radiographs, our approach achieved high accuracy in detecting bone loss severity (intra-class correlation coefficient up to $0.80$) and bone loss pattern classification (accuracy $87$\%). This automated system offers a rapid, objective, and reproducible tool for periodontal assessment, reducing reliance on subjective manual evaluation. By integrating AI into dental radiographic analysis, our framework has the potential to improve early diagnosis and personalized treatment planning for periodontitis, ultimately enhancing patient care and clinical outcomes.

\end{abstract}
\begin{document}

\flushbottom
\maketitle
%
%
\thispagestyle{empty}


\section*{Introduction}


The periodontium plays a crucial role in maintaining tooth health by providing structural support, protection, and facilitating the functional integration of teeth within the oral cavity. It consists of four main components: the gingiva, periodontal ligament, cementum, and alveolar bone, each contributing to the overall health and stability of teeth~\cite{melcher1976repair}. 
The health of the periodontium could be affected by several diseases and conditions. The most common disease, periodontitis is a chronic immuno-inflammatory condition which is primarily caused by bacterial infection that leads to progressive destruction of the alveolar bone and periodontal ligament, ultimately resulting in tooth loss if untreated~\cite{milward2019assessing}. 

Severe periodontitis has a global prevalence of about 19\% in individuals over 15 years of age and is also the sixth most prevalent disease worldwide~\cite{world2022global}. Smoking and uncontrolled diabetes have been identified as risk factors for exacerbation of the disease~\cite{matthews2007hyperactivity}. Further, it has a major impact on patients’ self-esteem since it could eventually lead to tooth loss, which compromises aesthetics and dietary intake~\cite{kaur2017impact}. 

The immuno-inflammatory reactions that occur due to the dysbiosis of the oral biofilm trigger a cascade of reactions that would affect periodontal structures, giving rise to various clinical and radiological presentations~\cite{matthews2007hyperactivity}. Appropriate evaluation of these clinical and radiographic findings during the patient assessment assists the clinician to arrive at a diagnosis as well as to determine the appropriate treatment plan~\cite{milward2019assessing}. Radiological assessments play a pivotal role in periodontal assessment~\cite{corbet2009radiographs}. The most frequently employed techniques include obtaining intra-oral radiographs, including peri-apical and bitewing radiographs~\cite{scarfe2017emerging}. 

Depending on the probing depths detected, for the sites with deep periodontal pockets, periapical radiographs are usually recommended, while for mild-moderate pocket depths, a bitewing radiograph could be applicable~\cite{jacobs2024radiographic}.
The identification of alveolar bone loss is critical in periodontal diagnosis, which would reflect as the clinical attachment loss~\cite{bragger2005radiographic}. Therefore, during radiological assessment, the extent of bone loss is assessed aligned with the long axis of the neighboring teeth, by evaluating the vertical distance between the \gls{cej} and the alveolar crest~\cite{hausmann1992computerized}. Figure~\ref{fig1_fig2_fig3_fig4:1} illustrates the key points used to assess alveolar bone loss on an \gls{iopa}.

\input{figures/fig1_fig2_fig3_fig4_TEX}

This reduction in bone height also causes changes in the alveolar crestal bone morphology. Understanding the nature of these alterations is essential for effective treatment planning. Radiologically, two main alveolar bone loss patterns could be identified as horizontal or angular (vertical) bone loss~\cite{esmaeli2012determination}. The horizontal pattern shows a uniform reduction in alveolar bone height around the teeth, as seen in Figure~\ref{fig1_fig2_fig3_fig4:3}, while the vertical (angular) pattern shows an uneven reduction that creates an angular appearance~\cite{moran2021radiographic}, as shown in Figure~\ref{fig1_fig2_fig3_fig4:4}. Clinicians use radiographic analysis to differentiate between these two bone loss patterns, which would assist in treatment planning, especially for periodontal regenerative approaches. 

Periodontal disease management is a complex and time-consuming process requiring long-term maintenance following treatment~\cite{tonetti2017impact}. Early diagnosis and prompt treatment significantly affect the outcome of treatment~\cite{shaddox2010treating}. 
At present, \gls{ai} has emerged as a transformative force, reshaping nearly every aspect of human life~\cite{gruetzemacher2022transformative, pachegowda2023globalimpactaiartificialintelligence, hagerty2019global}. \Gls{ai} enables machines to learn from as well as to overcome obstacles by taking inspiration from human cognition, especially the human's capacity for problem-solving~\cite{sen2020artificial, mccarthy2006proposal}. Computer vision~\cite{schwab2020localization, pomponiu2016deepmole, esteva2017dermatologist, levine2019rise}, machine learning~\cite{erickson2017machine, giger2018machine, zhang2014machine}, and \gls{dl}~\cite{litjens2017survey, shen2017deep, suganyadevi2022review, lee2017deep} are increasingly used in medical image analysis. \Gls{ai} could detect complex information and help to select more suitable treatment options by analyzing individual patient data and medical records~\cite{shiwlani2024revolutionizing}. They could also simplify complex processes, solve healthcare problems, and improve patient outcomes around the world~\cite{ahmed2021artificial}. According to \gls{dl}-based studies, which were carried out using 28 \gls{dl}-based articles in different fields of dentistry, AI models have shown promising results in dentistry~\cite{farook2021machine, corbella2021applications}.

Traditionally, the dentist will identify important anatomical sites and crestal morphology, denoting the bone loss pattern on the radiograph during radiological assessment. However, such assessments are highly subjective and time-consuming, frequently resulting in inconsistent diagnoses~\cite{arefin2024role}.

Recently, the use of AI has been suggested as a supportive tool for clinicians in diagnosing periodontitis and automatically identifying alveolar bone loss~\cite{revilla2023artificial}. These methods could minimize human error and streamline data analysis as well as task execution~\cite{revilla2023artificial}. Table~\ref{literature_review_overview} summarizes dataset information, study focus, and key findings of previous studies that had explored various approaches assessing periodontal bone loss. It is worth noting that, although these studies are useful and informative, they also have notable limitations.

\input{tables/literature_review_overview}

Several previous studies~\cite{bayrakdar2020success, sunnetci2022periodontal, alotaibi2022artificial} have classified full dental radiographs using binary or multi-class approaches. However, these methods do not provide detailed, tooth-specific, or region-specific information on bone loss, which is essential for accurate periodontal assessment. Several previous studies have used a detection method~\cite{ryu2023automated} and a segmentation method~\cite{uzun2023assessing, kurt2024detection} to identify the bone loss regions. However, these methods also failed to offer a detailed analysis of bone loss at the individual tooth level. Chang et al.~\cite{chang2020deep} used panoramic radiographs to calculate bone loss severity, while Tsoromokos et al.~\cite{tsoromokos2022estimation}, Lee et al.~\cite{lee2022use}, and Chen et al.~\cite{chen2023automatic} used peri-apical radiographs to calculate bone loss severity. Tsoromokos et al.~\cite{tsoromokos2022estimation} assessed radiographs of only mandibular teeth in the analysis. Other types of teeth were excluded to avoid potential data inconsistencies. Kurt-Bayrakdar et al.~\cite{kurt2024detection} study elaborates a technique to find the pattern of bone loss in an interdental region defined by the imaginary lines passing through the \gls{cej} points of two adjacent teeth and the alveolar bone crests. If these lines were parallel to each other, bone loss was considered horizontal, whereas if the lines were angular, it was considered vertical. However, the accurate applicability of this method in peri-apical radiographs was limited. Therefore, an improved method is necessary to correctly identify these cases.



Moreover, current studies do not analyze both the severity and patterns of alveolar bone loss using peri-apical radiographs. It is important to identify the alveolar bone loss pattern along with bone loss severity in order to select the correct and effective treatment plan~\cite{jayakumar2010horizontal}. Understanding the pattern of bone loss aids in planning appropriate interventions, aiming to manage and mitigate the effects of periodontal disease more effectively. To address this gap, we propose a novel AI-based framework for automated assessment of alveolar bone loss severity and patterns using IOPA radiographs. This is very useful for early detection and personalized treatment strategies in periodontal disease management. Our study makes the following contributions:

\begin{itemize}
\item We present a novel deep-learning pipeline capable of automating the detection and quantification of both the severity and pattern of alveolar bone loss. To the best of our knowledge, this study is the first to apply \gls{dl} techniques for the simultaneous prediction of these two critical aspects of alveolar bone loss.
\item Our method was evaluated on a large, expertly annotated dataset, and the results demonstrate the effectiveness and validity of our novel approach.
\item This study was designed through a collaboration between AI experts and dental professionals to ensure both the technical performance and practical applicability of the findings.
\end{itemize}

\section*{Materials and methods}


\subsection*{Dataset overview}
The DenPAR dataset~\cite{wimalasiri_2024_14181645}, consisting of $1000$ \gls{iopa} radiograph images, was used in this research. The dataset includes detailed annotations such as tooth segmentation masks,\gls{iopa} points, apex points, and alveolar crestal bone levels, all of which were annotated using the Labelbox tool. The images are accompanied by metadata on age and sex, and the dataset was split into three sets: training ($65$\%), validation ($15$\%), and testing ($20$\%). These splits were determined based on a statistical analysis that ensured balanced representation across important attributes, including the number of teeth in each radiograph, the arch type (mandibular or maxillary), and the presence of key anatomical landmarks like \gls{cej} and apex points. 

\subsection*{Calculate alveolar bone loss severity}

To identify alveolar bone loss, three specific points needed to be located: i) \gls{cej}, ii) alveolar bone level and tooth intersections, and iii) root apex, as shown in Figure~\ref{fig1_fig2_fig3_fig4:1}. First, YOLOv8x was used to detect the rectangular bounding box coordinates of each tooth in the radiographic images. Next, these rectangular bounding box coordinates, along with the radiographic images, were passed to three separate Keypoint R-CNN models. Each model was designed to predict the above-mentioned main types of key points for each tooth. Finally, all the predicted key points were combined to calculate the alveolar bone loss severity. This proposed system is illustrated in Figure~\ref{fig6_and_fig10:1}. This diagram illustrates the complete architecture of our system, highlighting key sections such as teeth detection and keypoint detection, combining the separate detected key points, and calculating the alveolar bone loss and its interactions.

\input{figures/fig6_and_fig10_TEX}

\subsubsection*{Implementation of the \gls{dl} pipeline}

You Only Look Once version 8 (YOLOv8)~\cite{reis2023real}, is a \gls{dl} model for real-time object detection. YOLOv8 is used in computer vision and \gls{dl} applications by using a single neural network to predict multiple bounding boxes and class probabilities simultaneously. The YOLOv8x is a variant of YOLOv8 that was used in our proposed system to identify the rectangular bounding box coordinates of each tooth. This approach was necessary because our keypoint detection models were trained on a per-tooth basis. Consequently, three distinct models were used to detect different types of keypoints. In each iteration, a tooth was processed individually through these models to predict the locations of the keypoints.  

Three Keypoint R-CNN~\cite{he2017mask} networks are used to identify these main types of key points separately from radiographs. Keypoint R-CNN, a type of Region-based Convolutional Neural Network, uses the Region of Interest (RoI) Align instead of RoI Pooling to mitigate the misalignment issues that arise due to quantization. This network extends the Faster R-CNN by adding a keypoint head, which is designed to predict the key points associated with detected objects. The outputs of this keypoint head are heatmaps for each keypoint of interest within the object-bounding boxes.

Keypoint R-CNN uses a backbone network to extract features from the input image. In our approach, we used ResNet-50~\cite{he2016deep} and incorporated another efficient network called a \gls{fpn}~\cite{lin2017feature} on top of it. Our backbone network extracts convolutional features from the entire input image at multiple scales. This is because ResNet-50 extracts features of the input image at varying levels of abstraction. \gls{fpn} further enhances these features using feature pyramids that capture information at different spatial resolutions. Hence, ResNet-50 with \gls{fpn} is the backbone network in our proposed model for keypoint detection.

After the main types of key points are detected separately, the outputs need to be combined to calculate alveolar bone loss. \gls{nms}~\cite{hosang2017learning} was used to achieve this task, using an \gls{iou} threshold of $0.6$ to filter overlapping key points. \gls{nms} is a technique used in object detection tasks, particularly in scenarios where multiple bounding boxes are detected for the same object instance. The primary purpose of \gls{nms} is to reduce redundant detections by selecting the most confident bounding boxes that correspond to the same object.

\subsubsection*{Calculating the alveolar bone loss severity using the mathematical approach}

Equation~\eqref{eq1} was used for computing alveolar bone loss severity, where a ratio between the pixel distances between points was used. Three points must align along the same line to get the ratio. To achieve this alignment, the min-max line~\cite{cook_minmax_2023} was used, minimizing the maximum error. 

\vspace{10mm}
\input{equations/eq1}

Under this approach, assuming our three points are denoted as $(a_{1}, b_{1})$, $(a_{2}, b_{2})$, and $(a_{3}, b_{3})$, with  $(a_{1} < a_{2} < a_{3})$. Then, the gradient of the line is,

\input{equations/eq2}

And the intercept of the line is,

\input{equations/eq3}

Figure~\ref{fig5} illustrates that after the min-max line was established, points can be projected onto it perpendicularly. Subsequently, the pixel distance between these points, necessary for determining alveolar bone loss severity, can be measured. This approach allows for the measurement of alveolar bone loss in teeth on both sides. 

\input{figures/fig5_TEX}

\subsubsection*{Experimental parameter setting}
For the tooth detection task, the YOLOv8x used the Adam~\cite{kingma2014adam} optimizer and a batch size of $4$. The initial learning rates were set to $0.0001$ for the model. Additionally, a cosine learning rate scheduler was used. To ensure the best performance of the models, an early stopping method was implemented with a patience size of $25$. The early stopping mechanism further enhanced model performance by preventing overfitting, ensuring that the models were saved at their optimal state.

For the keypoint detection task, Keypoint R-CNN models were set with an initial learning rate of $0.0001$ for detecting the \gls{cej}, the intersection point of the tooth boundary and the alveolar bone level, and the root apex. These models used the StepLR learning rate scheduler with a gamma value of $0.6$ and a step size of $4$. All models were trained using the Adam optimizer with a batch size of $8$. To ensure the best performance of the models, an early stopping method was implemented with a patience size of $30$.


\subsection*{Finding alveolar bone loss pattern}
To determine the bone loss pattern in a detected bone loss case, the bone level of the affected region must first be identified. In our approach, YOLOv8x-seg was used to identify the bone levels at affected sites, along with the tooth masks of the relevant teeth. An example of the bone level representation is shown in Figure~\ref{fig1_fig2_fig3_fig4:2}. Then, tooth masks were converted into polygons, representing the tooth outlines. Subsequently, by drawing two tangents at the intersection point of the tooth boundary and alveolar bone level, the bone loss angle was calculated. Using the reference values related to this bone loss angle, the pattern was finally determined. An overview of the proposed system is illustrated in Figure~\ref{fig6_and_fig10:2}. This diagram shows the complete architecture of the proposed system, highlighting key components such as bone-level line prediction, tooth mask prediction, and the geometrical approach used for bone loss pattern identification.

\subsubsection*{Implementation of the \gls{dl} models}
In a single radiograph, there can be multiple regions affected by alveolar bone loss. To detect the bone loss patterns in these regions, all the relevant bone level lines must be identified. In this case, an instance-segmentation approach was followed. This involved training a YOLOv8x-seg model using the bone-level segmentation masks as inputs.

YOLOv8x-seg is a variant of the YOLOv8, designed specifically for the dual tasks of object detection and instance segmentation. One of the important features of YOLOv8x-seg is its ability to perform instance segmentation. This allows for the accurate identification of object boundaries within images. This is very useful for applications requiring detailed object identification. 

Since the dataset only contained alveolar bone levels represented as lines, a data pre-processing method was needed to convert these bone lines into segmentation masks. On the left side of Figure~\ref{fig7_and_fig8:1} is the binary mask created using the bone line annotations as they are. By applying the pre-processing method, the thickness of the bone lines was increased to $10$ pixels, creating a more visible set of binary masks without losing the basic shape of the bone level line, as shown on the right side of Figure~\ref{fig7_and_fig8:1}. All these bone lines are drawn only in the gaps between the teeth.

On the left side of Figure~\ref{fig7_and_fig8:2} is a set of bone-level segmentation masks predicted by the \gls{dl} model. To perform the bone loss pattern-finding calculation, these segmentation masks needed to be converted into lines. A method was used to extract the central path through each mask along its longest extent, resulting in the desired line representation.

\input{figures/fig7_and_fig8_TEX}

The method for this conversion begins by identifying the leftmost point on the mask boundary. Each point before and after this initial point is paired sequentially, continuing in this manner with subsequent outer points on both sides (for example, pairing the second point before with the second point after, the third point before with the third point after, etc.), as indicated by the dotted lines in subfigure (i) of Figure~\ref{mask_to_line_and_fig9:1}. The midpoint of each pair is then determined, and these midpoints are connected to form a new line, represented as a red line in subfigure (i) of Figure~\ref{mask_to_line_and_fig9:1}.

To ensure a more balanced approach, the same procedure is repeated starting from the rightmost point, resulting in a second line as shown in subfigure (ii) of Figure~\ref{mask_to_line_and_fig9:1}. An average line is then created by calculating the mean of the corresponding points on both lines, as illustrated in subfigure (iii) of Figure~\ref{mask_to_line_and_fig9:1}, which yields a final line similar to the green line represented in subfigure (iv) of Figure~\ref{mask_to_line_and_fig9:1}. This method is applied to all the bone segmentation masks identified in the dental radiographs, thereby transforming them into bone-level lines, as shown on the right side of Figure~\ref{fig7_and_fig8:2}.

In addition to identifying the bone level lines, the tooth boundary needed to be identified to determine the bone loss pattern, as it is defined relative to the respective tooth face. To identify the tooth outline, YOLOv8x-seg was also used. The tooth masks obtained through the YOLOv8x-seg were converted into polygons, representing the tooth outlines.

\input{figures/mask_to_line_and_fig9_TEX}

\subsubsection*{Determining the bone loss pattern using the mathematical approach}

After identifying the bone level lines and the tooth outlines, a geometrical method was applied to determine the bone loss pattern. This was carried out at the locations where the bone loss was detected via the bone loss severity calculations. As represented in Figure~\ref{mask_to_line_and_fig9:2}, first, a tangent line (yellow) was drawn to the predicted bone line (blue) at the intersection point of the predicted tooth boundary and the alveolar bone level.

Next, a tangent line to the tooth face was drawn, as shown in red in Figure~\ref{mask_to_line_and_fig9:2}. Similar to the bone line tangent, this line was drawn at the intersection point of the predicted tooth boundary and the alveolar bone level.

After drawing the two tangent lines, the unit vectors along the two lines were calculated. The unit vector along the tooth face tangent was always taken in such a way that it would be in the direction from the tooth crest to the tooth crown. The unit vector along the tangent drawn to the bone line was taken in such a way that it would always direct away from the tooth face.

After finding the two unit vectors, the vector dot product was utilized to determine the angle between the two vectors, as shown in~\eqref{eq4}, which could be approximately used as an indicator for the bone loss pattern. In cases where this angle~$\theta$ was less than $54.1372^\circ$, the instance was identified as an angular bone loss case. This value was identified by doing a statistical analysis with the $720$ ground-truth annotations given by the dental experts. Figure~\ref{fig20} shows the angle values of angular and horizontal cases and also the $54.1372^\circ$ threshold horizontal line. If the angle was greater than $54.1372^\circ$, it was identified as a horizontal bone loss case. The same geometrical method was applied to all detected bone loss cases to determine bone loss patterns.

\input{equations/eq4}

where $\theta$ is the angle between two vectors, and u and v are unit vectors taken from the two tangents.

\input{figures/fig20_TEX}

\subsubsection*{Experimental parameter setting}
To individually identify each tooth and bone line mask, YOLOv8x-Seg used the Adam optimizer with a batch size of $4$. The initial learning rate was set to $0.0001$, and a cosine learning rate scheduler was applied. Additionally, an early stopping method with a patience of $30$ epochs was implemented to ensure optimal performance.

\section*{Results and Discussions}

\subsection*{Calculate alveolar bone loss severity}

First, the detection of the bounding box of each tooth was experimented with using three models: i) YOLOv8n, ii) YOLOv8x, and iii) YOLOv9e. These models are different variants of YOLOv8 and YOLOv9~\cite{wang2024yolov9learningwantlearn}. The results for each model are detailed in Table~\ref {tooth_detection_results}. All three models were trained under the same conditions for each parameter, including initial learning rate, batch size, optimizer, learning rate scheduler, and patience. The results for each model are detailed in Table~\ref{tooth_detection_results}.

\input{tables/tooth_detection_results}

\input{figures/fig11_TEX}

The test set is considered unseen data for the model during training. Therefore, test results can be used to select the best model. As shown in Table~\ref{tooth_detection_results}, YOLOv8x shows the highest \gls{map} and \gls{map50}, indicating strong performance in detecting objects with varying degrees of overlap. YOLOv8n has the highest precision among the models, and also \gls{map50} is similar to YOLOv8x. YOLOv9e has a balanced performance but does not lead in any metric. \gls{map50} is a more comprehensive metric reflecting the overall accuracy across different thresholds. Therefore, \textbf{YOLOv8x} appears to be the best model for this tooth detection task. Figure~\ref{fig11} illustrates the ground truth and predicted results of YOLOv8x for this task, and the model could be able to identify the teeth with the teeth class and higher confidence results.

Keypoint R-CNN~\cite{he2017mask} and YOLOv8x-pose~\cite{maji2022yolo} were experimented to detect main types of keypoints. \gls{oks} was used as the threshold to evaluate the key points. \gls{oks} is considered the \gls{iou} in keypoints~\cite{maji2022yolo}. By taking into account variables like scale, unlabelled keypoints, and annotation ambiguity, \gls{oks} offers a standardized method for comparing the predicted keypoints with the ground truth~\cite{learnopencvObjectKeypoint}. 

\input{tables/keypoint_results}

\input{figures/fig12_and_fig13_and_fig14_TEX}

In this experiment, two models were trained using identical parameter settings, with the only differences being the initial learning rate and the learning rate scheduler to achieve the best possible results. For Keypoint R-CNN, the initial learning rate was set to $0.0001$, and a StepLR scheduler with a gamma value of $0.6$ and a step size of $4$ was used. For YOLOv8 pose, the initial learning rate was $0.00001$, and a Cosine learning rate scheduler was used. The results of the model comparisons are presented in Table~\ref{keypoint_results}.

As shown in Table~\ref{keypoint_results}, when comparing test results, the \textbf{Keypoint R-CNN} model performed better than the YOLOv8 Pose model in detecting all types of keypoints. Figure~\ref{fig12_and_fig13_and_fig14:1} illustrates the output results with ground truth produced by the Keypoint R-CNN for \gls{cej}, Figure~\ref{fig12_and_fig13_and_fig14:2} for the intersection of tooth and bone line, and Figure~\ref{fig12_and_fig13_and_fig14:3} for apex points.

After combining all the keypoints, we used our mathematical approach for ground truth keypoints and found the ground truth bone loss severity. Then, we compared those values with our system output severity. \gls{icc} was used for evaluation. The \gls{icc}s of training, validation, and testing are $0.851$, $0.824$, and $0.801$, respectively. All \gls{icc} values are between $0.75$ and $0.90$, which indicates good reliability according to the guidelines provided by Koo and Li (2016)~\cite{koo2016guideline}. Figure~\ref{fig15} demonstrates the calculation of alveolar bone loss severity using the proposed mathematical approach, showing the severity results from both ground truth points and predicted points. Additionally, it includes the best-fit line, keypoints, and the severity of alveolar bone loss. This finding suggests that the proposed mathematical approach is effective in estimating the severity of alveolar bone loss. 

\input{figures/fig15_TEX}

\input{tables/combined_mask_prediction_results}

\input{figures/fig16_and_fig17_TEX}

\subsection*{Calculate Alveolar Bone Loss Pattern}

To calculate the alveolar bone loss pattern, detecting both tooth boundaries and bone lines was required. For that, we had to identify each tooth individually. For both cases, two models were experimented with: YOLOv8x-Seg, which is a variant of YOLOv8, and Mask R-CNN~\cite{he2017mask}. 

Both the YOLOv8x-Seg and Mask R-CNN models were trained using the Adam optimizer, but different parameter settings were applied to each model to optimize their performance. YOLOv8x-Seg was trained with a batch size of $4$, an initial learning rate of $0.0001$, and a cosine learning rate scheduler. In contrast, Mask R-CNN was trained with a batch size of $1$ and an initial learning rate of $0.00001$, using a Step learning rate scheduler with a step size of $10$ and a gamma of $0.9$. For both models, early stopping with a patience of $30$ epochs was used to prevent overfitting. These differences in training parameter settings were chosen based on each model to achieve the best possible results. The performance comparisons are presented in Tables~\ref{tooth_detection_results} and \ref{combined_mask_prediction_results}.

As shown in Table~\ref{combined_mask_prediction_results}, the YOLOv8x-seg model demonstrates better performance in predicting tooth masks compared to the Mask R-CNN model, based on the results from the test set. The \gls{ap50} value for YOLOv8x-seg is $0.978$ on the test set, outperforming the Mask R-CNN, which has an \gls{ap50} value of 0.900. The \gls{ap} for YOLOv8x-seg was also higher, with a value of $0.900$ on the test set, compared to Mask R-CNN's $0.548$. These results indicate that \textbf{YOLOv8x-seg} is more effective for tooth mask prediction, making it the preferred model for this task. After that, tooth boundaries were taken from the predicted tooth masks. Figure~\ref{fig16_and_fig17:1} illustrates ground truth and output results produced by YOLOv8x-seg for tooth boundary detection.

As shown in Table~\ref{combined_mask_prediction_results}, based on the results obtained from the test set, the YOLOv8x-seg model demonstrates good performance in predicting the teeth mask compared to the Mask R-CNN model. The \gls{ap50} value for YOLOv8x-seg is $0.525$ on the test set, outperforming the Mask R-CNN, which has an \gls{ap50} value of $0.005$. The \gls{ap} for YOLOv8x-seg is higher, with a value of $0.135$ on the test set, compared to Mask R-CNN's $0.001$. These results indicate that \textbf{YOLOv8x-seg} is effective for bone line mask prediction, making it the preferred model for this task. The bone-level masks taken through this model were further converted into bone-level lines as mentioned in the methodology section. Figure~\ref{fig16_and_fig17:2} illustrates the output results with ground truth produced by YOLOv8x-seg for bone line mask detection.

\input{tables/comparison_various_pixel_levels}

\input{figures/fig21_TEX}

As mentioned in the methodology section, the bone level lines were preprocessed and converted into masks by setting a thickness value for the bone line. The experiments were conducted with a range of thickness values (in pixels) to find the optimal thickness value that yields the best bone line mask detection. For the experiment, the model was used to train, validate, and test several pixel-level masks and convert those masks to lines. After that, predicted bone lines were compared with ground truth lines. Table~\ref{comparison_various_pixel_levels} shows the \gls{mse} between the ground truth bone lines and predicted bone lines generated from different pixel levels.    

Considering the test errors, a thickness value of $10$ pixels provides consistently good performance across all datasets for the bone line detection task, based on the test \gls{mse}s. We used $720$ ground truth annotations of bone loss patterns, all independently assessed and verified by experienced dental experts as described in the methodology section. These expert assessments served as the reference standard for comparing our model’s predictions. Out of the $720$ annotated cases, $81$ were excluded—mainly due to missing bone line or teeth masks in the model’s output—leaving $639$ cases for evaluation. We then used a confusion matrix as shown in Figure~\ref{fig21} to evaluate the model's performance, resulting in an accuracy of $0.869$, a precision of $0.985$, a recall of $0.826$, and a sensitivity of $0.872$. Our method could identify both horizontal and vertical cases most accurately.

Figure~\ref{fig18} displays the results after applying the mathematical calculations to the bone level line and tooth boundary results. The left column presents a set of ground truth values for bone loss cases marked by dental professionals, while the right column shows the bone loss cases identified by the proposed methodology. In both columns, red circles indicate angular bone loss, and other markings represent horizontal bone loss. The proposed methodology correctly identified the cases when compared to the ground truth.

\input{figures/fig18_TEX}

\section*{Conclusion}
In this study, we developed an AI-based method for automatically detecting bone loss patterns and measuring alveolar bone loss severity from \gls{iopa} radiographs. According to our results, our deep learning fine-tuned models provide precise and efficient analysis. Such a development can help dentists in the early diagnosis and specific treatment of periodontitis. With the help of deep learning methods and computer vision, our system is capable of detecting both the severity and patterns of bone loss, which are important for differential diagnosis and further treatment planning.

The integration of \gls{ai} into dental diagnostics is a significant step forward, enhancing the dynamic features of today’s healthcare systems. This helps to modernize healthcare practices. Traditional methods of diagnosing alveolar bone loss are mostly time-consuming and depend on the dentist's experience, making them subjective. This could lead to inconsistencies in diagnosis. On the other hand, the AI-based approach presents a standardized, objective, and rapid solution, which increases the reliability in dental assessments.

\section*{Author contributions statement}
S.W., C.W., and P.R. conducted the experiments and literature review, S.R. and I.N. conceptualized the study, I.N., V.T., and R.G.R. provided engineering expertise and resources, S.R. and D.L. provided dental expertise, I.N., S.R., V.T., D.L. and R.G.R provided supervision and reviewed methodologies. All authors contributed to writing and reviewed the manuscript. 

\section*{Competing interests}
The authors declare no competing interests.

\section*{Data Availability}
The dataset which this study is based on is available on Zenodo at \href{https://doi.org/10.5281/zenodo.14181645}{https://doi.org/10.5281/zenodo.14181645}. 

\section*{Code availability}
The deep learning models and algorithms used in our method are available at the linked \href{https://github.com/chathurawimalasiri/analysis-in-detecting-alveolar-bone-loss}{GitHub repository}.






 





\bibliography{references}

\end{document}

%% file: acronyms.tex
\newacronym{iopa}{IOPA}{intraoral periapical radiograph}
\newacronym{cej}{CEJ}{cemento-enamel junction}
\newacronym{ai}{AI}{artificial intelligence}
\newacronym{dl}{DL}{deep learning}
\newacronym{nms}{NMS}{Non-maximum suppression}
\newacronym{fpn}{FPN}{Feature Pyramid Network}
\newacronym{iou}{IoU}{Intersection over Union}
\newacronym{map}{mAP}{Mean Average Precision}
\newacronym{oks}{OKS}{Object Keypoint Similarity}
\newacronym{icc}{ICC}{Intra-class correlation coefficient}
\newacronym{ap50}{AP50}{average precision at the IoU threshold of $0.5$}
\newacronym{map50}{mAP50}{mean average precision at the IoU threshold of $0.5$}
\newacronym{mse}{MSE}{mean squared error}
\newacronym{ap}{AP}{Average Precision}

%% file: figures/fig1_fig2_fig3_fig4_TEX.tex
\begin{figure}[ht]
    \centering
    \begin{minipage}{0.7\textwidth}
        \centering
        \begin{subfigure}[t]{0.45\textwidth}
            \centering
            \includegraphics[width=\textwidth]{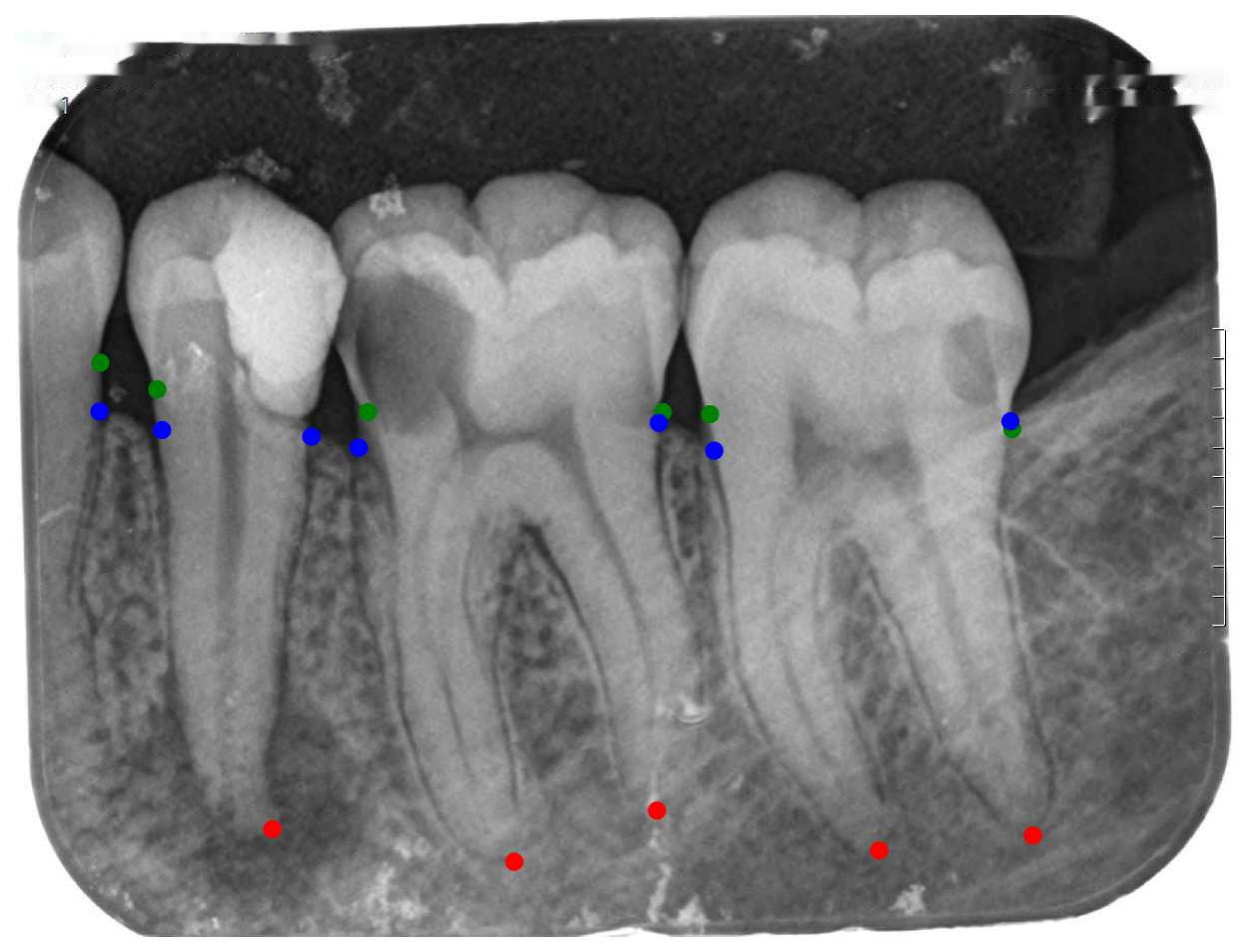}
            \caption{The key points required for alveolar bone loss analysis. Green points - \acrfullpl{cej}, blue points - the intersection points of the alveolar crest bone level and the tooth boundary, red points -  the apex points.}
            \label{fig1_fig2_fig3_fig4:1}
        \end{subfigure}
        \hfill
        \begin{subfigure}[t]{0.45\textwidth}
            \centering
            \includegraphics[width=\textwidth]{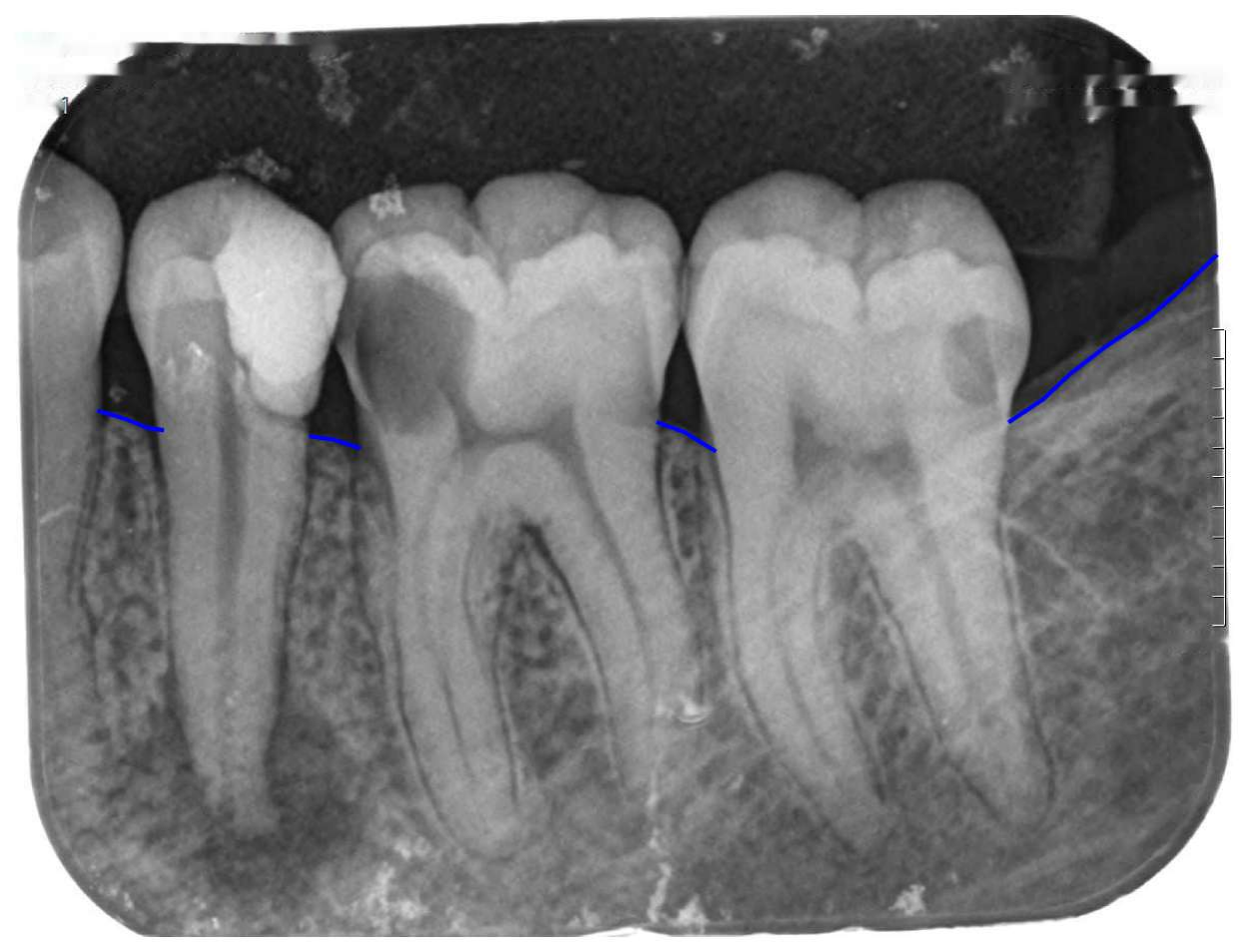}
            \caption{Alveolar bone levels required for alveolar bone loss pattern analysis, represented by blue lines.}
            \label{fig1_fig2_fig3_fig4:2}
        \end{subfigure}
    \end{minipage}

    \vspace{1em}

    \begin{minipage}{0.7\textwidth}
        \centering
        \begin{subfigure}[t]{0.45\textwidth}
            \centering 
            \includegraphics[height=4cm]{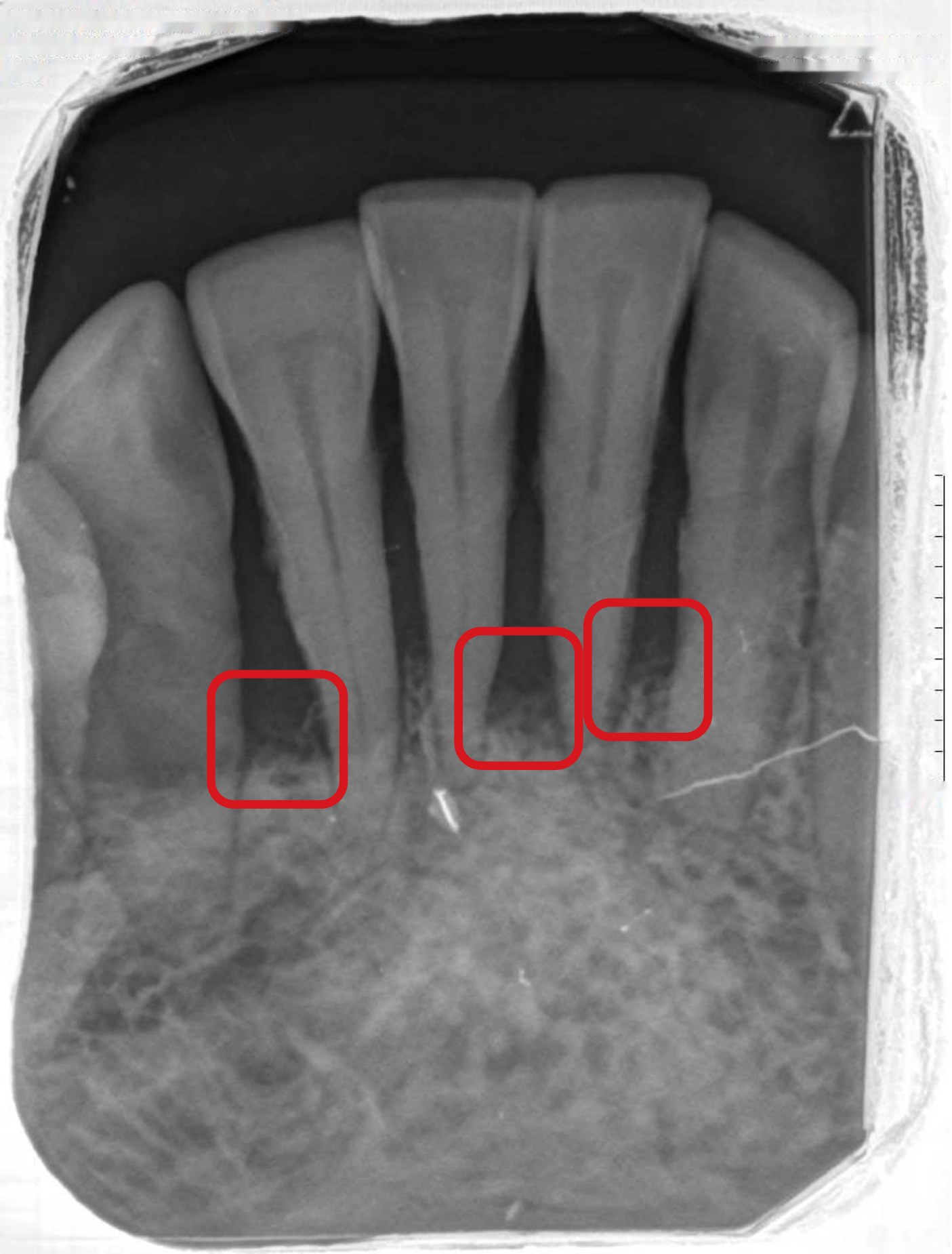}
            \caption{Cases of horizontal alveolar bone loss highlighted in red boxes, each enclosing two affected teeth.}
            \label{fig1_fig2_fig3_fig4:3}
        \end{subfigure}
        \hfill
        \begin{subfigure}[t]{0.45\textwidth}
            \centering
            \includegraphics[width=\textwidth]{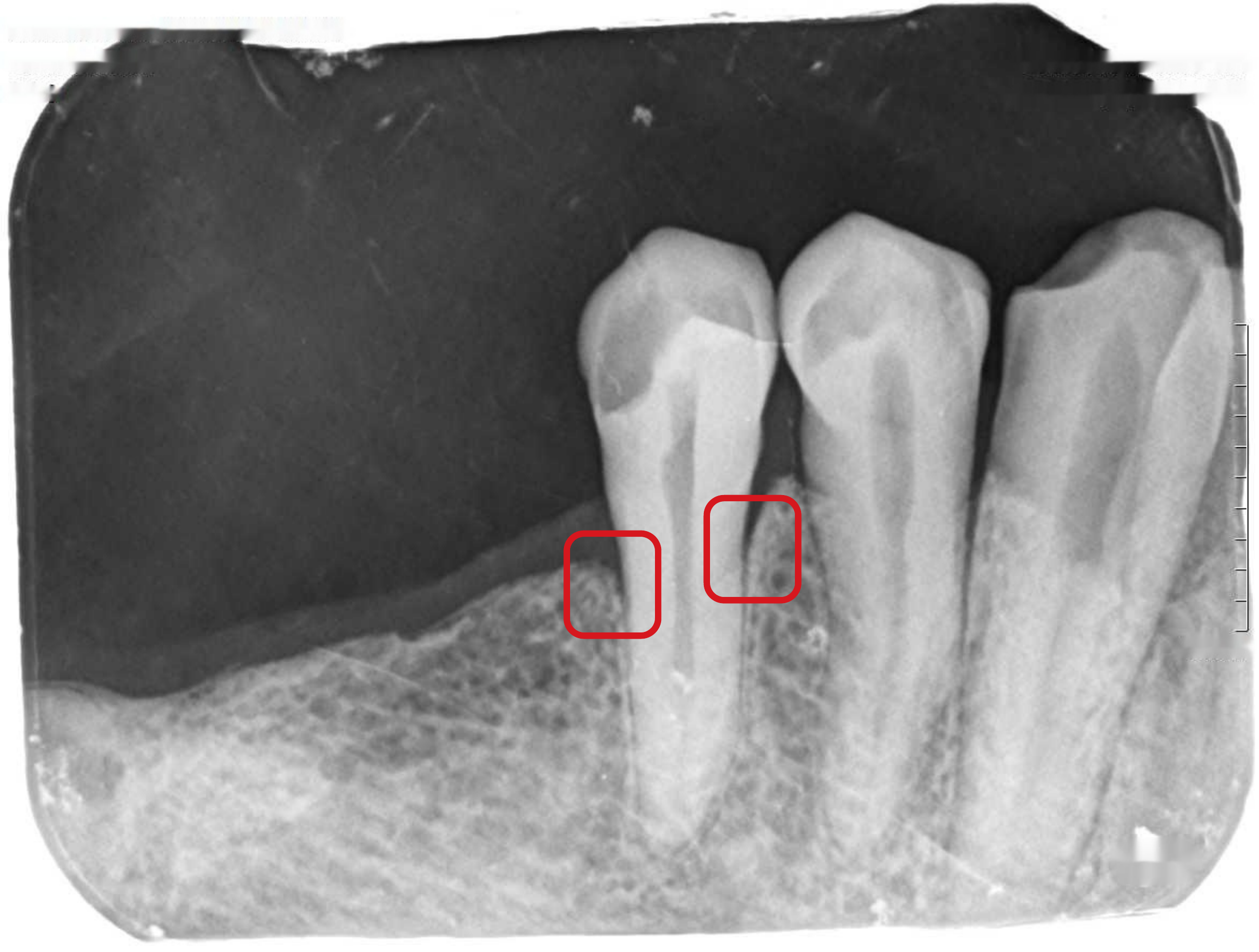}
            \caption{Cases of angular (vertical) alveolar bone loss highlighted in red boxes, each enclosing the affected side of a tooth.}
            \label{fig1_fig2_fig3_fig4:4}
        \end{subfigure}
    \end{minipage}

    \caption{Overview of key features and case classifications used in alveolar bone loss pattern analysis. (a) Key points required for alveolar bone loss analysis; (b) Alveolar bone levels for pattern assessment; (c) Alveolar bone loss cases with horizontal pattern; (d) Alveolar bone loss cases with angular (vertical) pattern.}
    \label{fig1_fig2_fig3_fig4}
\end{figure}

%% file: tables/literature_review_overview.tex
\begin{table}[ht]
\centering
\small
\begin{tabular}{|p{0.1\textwidth}|p{0.25\textwidth}|p{0.55\textwidth}|}
\hline
\textbf{Authors Name} & \textbf{Dataset Information} & \textbf{Study Focus and Key Findings} \\

\hline
Bayrakdar et al., 2020 \cite{bayrakdar2020success} &  Panoramic Radiographs: Healthy - $1139$, Bone Loss - $1137$ & Used convolutional neural network model (CNN) to classify healthy and bone loss cases.\\
\hline

Sunnetci et al., 2022 \cite{sunnetci2022periodontal} & Panoramic Radiographs: Healthy - $810$, Bone Loss - $622$ & Used a hybrid system combining deep learning and machine learning models to classify healthy and bone loss cases. Developed a user-friendly graphical interface. \\
\hline

Alotaibi et al., 2022 \cite{alotaibi2022artificial} &  Periapical Radiographs: Healthy - $814$, Bone Loss - $910$ (Mild - $511$, Moderate - $290$, Severe - $109$) & Used a CNN network to classify normal and severity of bone loss cases of the anterior region.\\
\hline

Ryu et al., 2023 \cite{ryu2023automated} & Panoramic Radiographs: $4083$ & Grouping teeth into anterior, premolar, and molar regions. Faster R-CNN was used to detect those regions as healthy, edentulous, and periodontitis. \\
\hline

Uzun Saylan et al., 2023 \cite{uzun2023assessing} & Panoramic Radiographs: $685$ labeled out of $1543$ & YOLO-v5 model detected alveolar bone loss regions of the jaw.\\
\hline

Chang et al., 2020 \cite{chang2020deep} & Panoramic Radiographs: $340$ & Used a hybrid method combining CNNs with traditional computer aided design (CAD) to classify each tooth into stages of periodontitis. The intraclass correlation (ICC) was $0.91$ between the proposed system and the radiologists’ diagnoses.\\
\hline

Tsoromokos et al., 2022 \cite{tsoromokos2022estimation} & Periapical Radiographs: $446$ & Developed a CNN model to detect and quantify alveolar bone loss (ABL). The ICC was $0.601$ across all teeth. The ICC was $0.763$ for nonmolar teeth.\\
\hline

Lee et al., 2022 \cite{lee2022use} & Periapical Radiographs: $693$ & The U-Net model architectures were used for segmentation tasks. Calculated the severity of radiographic bone loss (RBL) and assigned RBL stages based on the severity of RBL. The Kappa coefficient was $0.81$ between the proposed system and the ground truth of RBL stage assignment.\\
\hline

Chen et al., 2023 \cite{chen2023automatic} & Periapical Radiographs: $8000$  & CNN models were used to detect tooth position, detect shape, remaining interproximal bone level, and finally calculate RBL. Research does not fully report descriptive statistics (e.g., ICC, kappa) for comparing predicted and ground truth RBL values.\\
\hline

Kurt-Bayrakdar et al., 2024 \cite{kurt2024detection} & Panoramic Radiographs: $1121$ &  A CNN-based system was used to identify total alveolar bone loss region, horizontal bone loss, vertical bone loss, and furcation defect. Area under the ROC curve (AUC) was $0.910$, $0.733$ for horizontal and vertical bone loss cases, respectively.\\
\hline

\end{tabular}
\caption{\label{literature_review_overview}Overview of studies on the application of deep learning for bone loss detection in radiographs}
\end{table}

%% file: figures/fig6_and_fig10_TEX.tex
\begin{figure}[ht]
    \centering
    \begin{subfigure}[b]{1.0\textwidth}
        \centering
        \includegraphics[width=\textwidth]{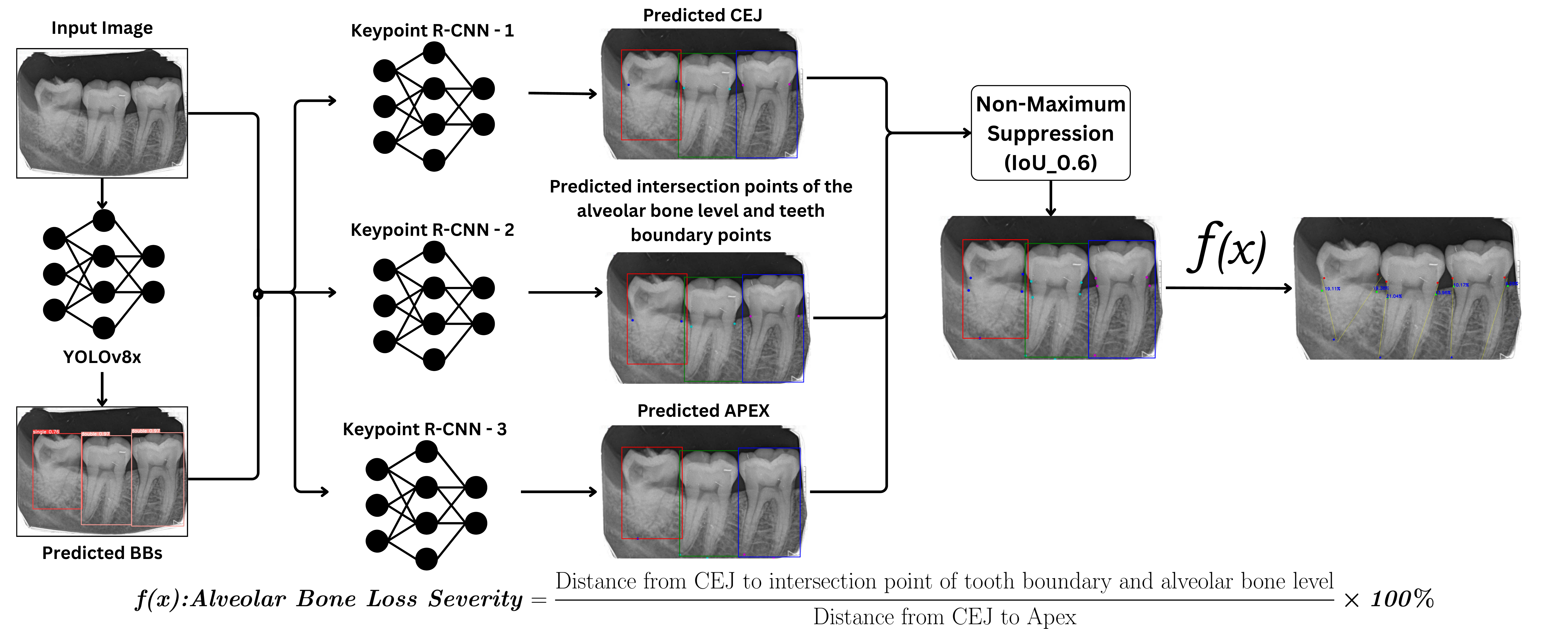}
        \caption{Overall architecture of the proposed system for calculating alveolar bone loss severity.}
        \label{fig6_and_fig10:1}
    \end{subfigure}
    
    \vspace{0.8em} 

    \begin{subfigure}[b]{1.0\textwidth}
        \centering
        \includegraphics[width=\textwidth]{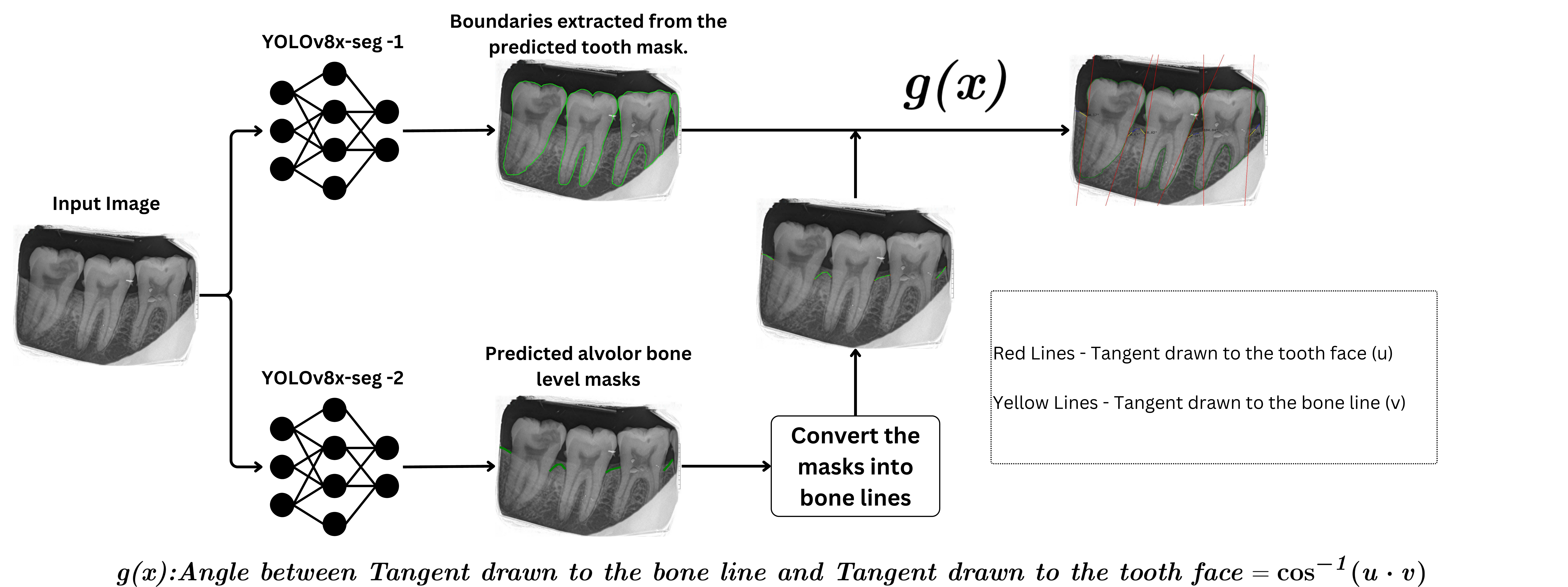}
        \caption{Overall architecture of the proposed system for finding alveolar bone loss pattern.}
        \label{fig6_and_fig10:2}
    \end{subfigure}
    \caption{Overview of the proposed system: (a) architecture for calculating alveolar bone loss severity; (b) architecture for identifying alveolar bone loss pattern.}
    \label{fig6_and_fig10}
\end{figure}

%% file: equations/eq1.tex
\begin{equation}
\text{Alveolar Bone Loss Severity} = \frac{\text{Distance from CEJ to intersection point of tooth boundary and alveolar bone level}}{\text{Distance from CEJ to Apex}} \times 100\%
\label{eq1}
\end{equation}

%% file: equations/eq2.tex
\begin{equation}
m = \frac{b_{3}-b_{1}}{a_{3}-a_{1}}\label{eq2}
\end{equation}

%% file: equations/eq3.tex
\begin{equation}
c = \frac{b_{1}(a_{2}+a_{3}) + b_{2}(a_{3}-a_{1}) - b_{3}(a_{1}+a_{2})}{2(a_{3}-a_{1})}\label{eq3}
\end{equation}

%% file: figures/fig5_TEX.tex
\begin{figure}[ht]
\centerline{\includegraphics[width=0.5\linewidth]{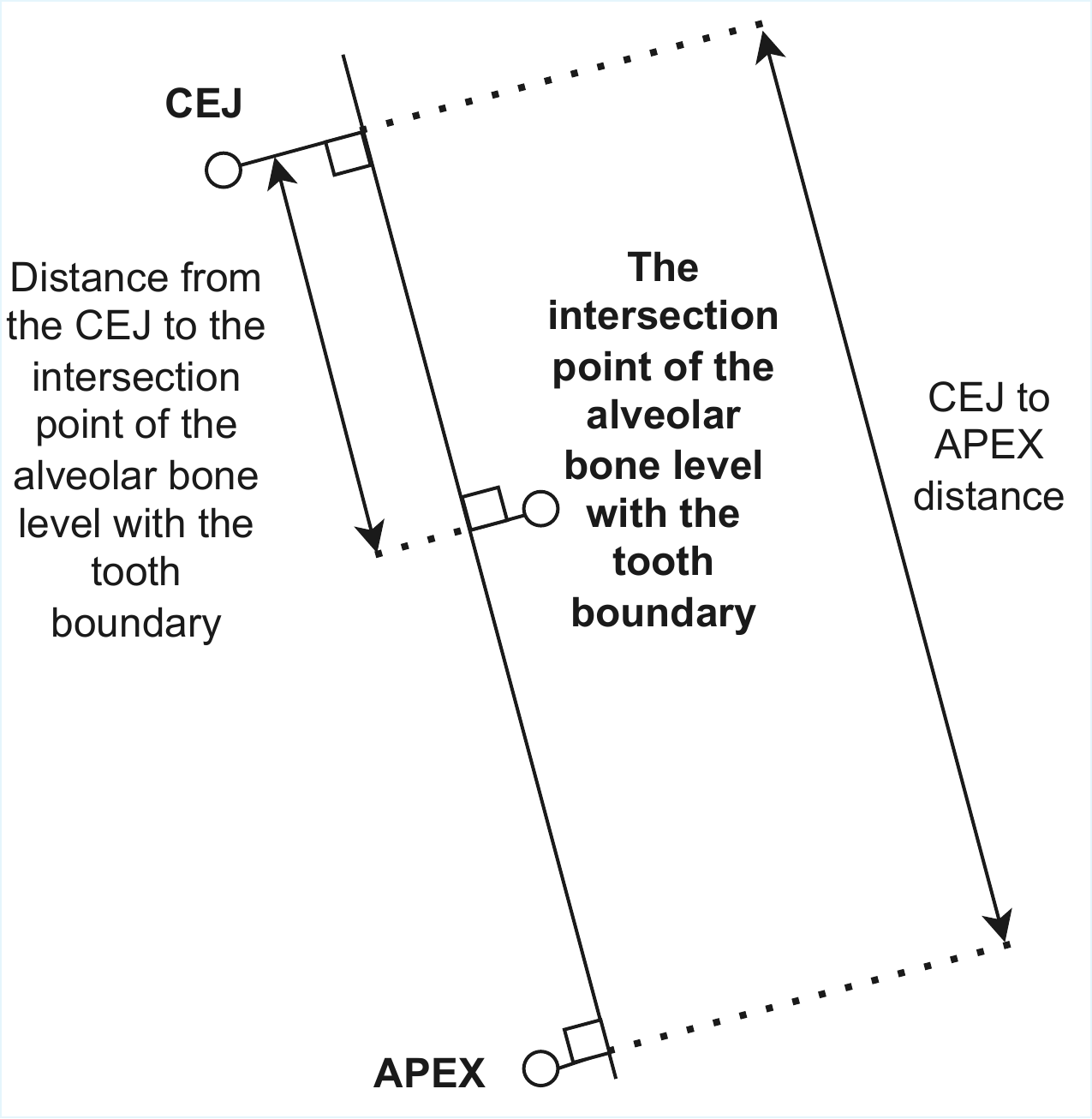}}
\caption{Projection of points onto a minimax line with perpendiculars.}
\label{fig5}
\end{figure}

%% file: figures/fig7_and_fig8_TEX.tex
\begin{figure}[ht]
    \centering
    \begin{subfigure}[t]{0.45\textwidth}
        \centering
        \includegraphics[width=\textwidth]{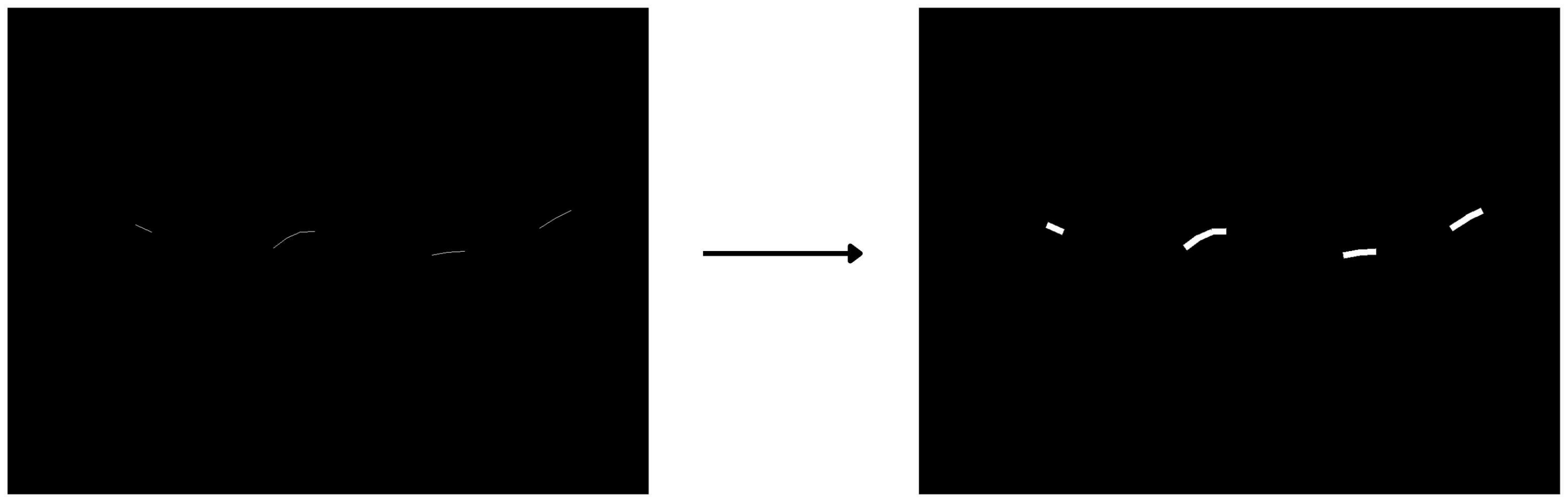}
        \caption{Pre-processing for converting lines into segmentation masks.}
        \label{fig7_and_fig8:1}
    \end{subfigure}
    \hfill
    \begin{subfigure}[t]{0.45\textwidth}
        \centering
        \includegraphics[width=\textwidth]{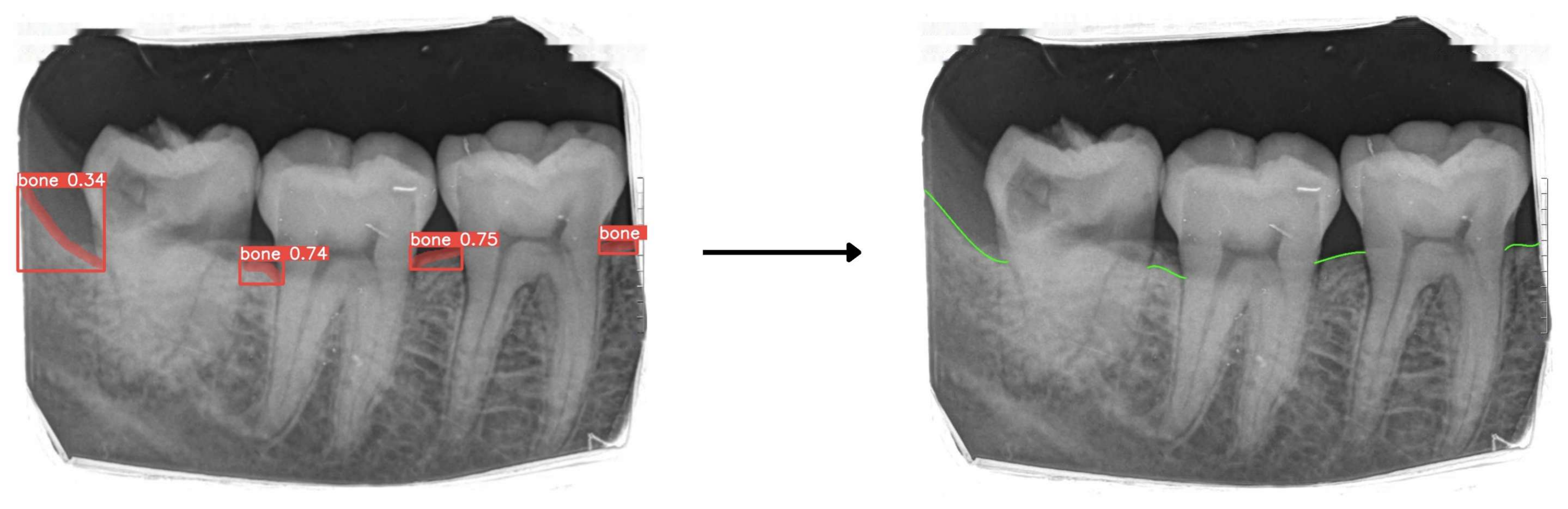}
        \caption{Post-processing for converting masks into line.}
        \label{fig7_and_fig8:2}
    \end{subfigure}
    \caption{Steps for converting between lines and segmentation masks: (a) Pre-processing lines into masks, and (b) Post-processing masks back into lines.}
    \label{fig7_and_fig8}
\end{figure}

%% file: figures/mask_to_line_and_fig9_TEX.tex
\begin{figure}[ht]
    \centering
    \begin{subfigure}[t]{0.65\textwidth}
        \centering
        \includegraphics[width=\textwidth]{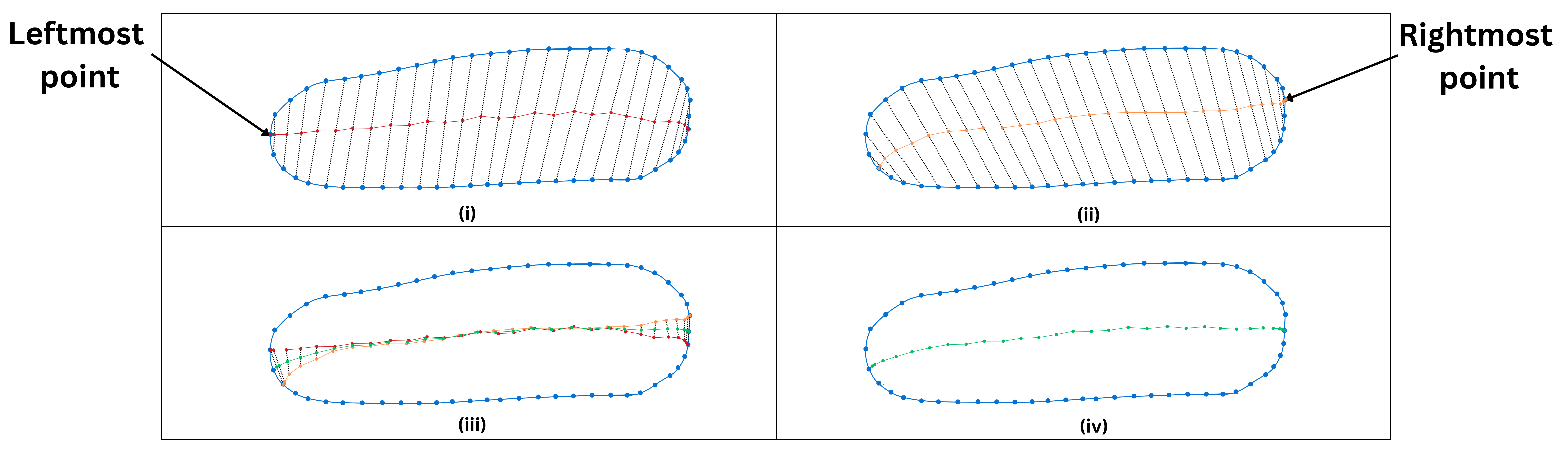}
        \caption{Method for converting bone level segmentation masks into bone level lines.}
        \label{mask_to_line_and_fig9:1}
    \end{subfigure}
    \hfill
    \begin{subfigure}[t]{0.30\textwidth}
        \centering
        \includegraphics[width=\textwidth]{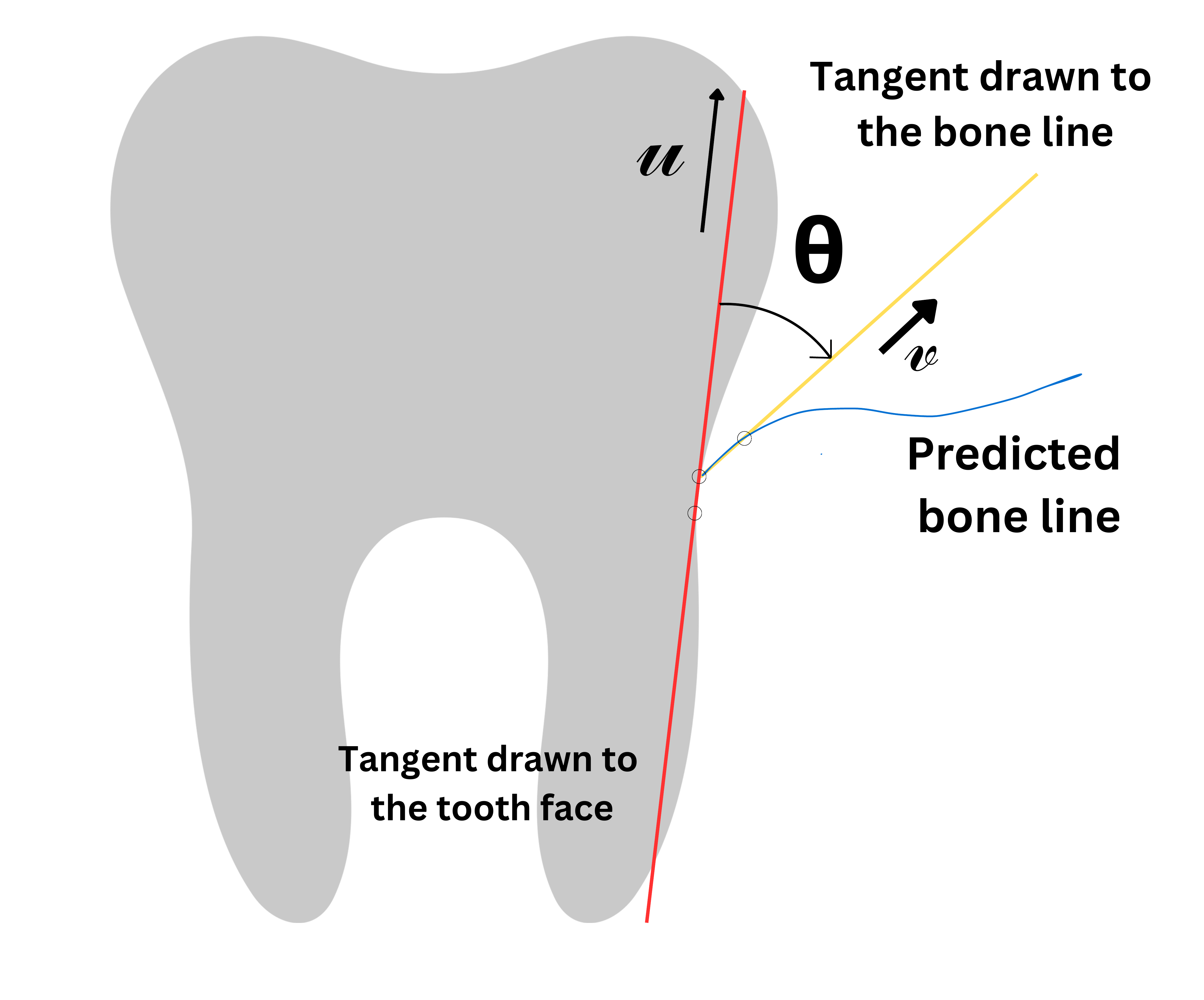}
        \caption{Geometrical method for calculating bone loss angle.}
        \label{mask_to_line_and_fig9:2}
    \end{subfigure}
    \caption{(a) Method for converting bone level segmentation masks into bone level lines. (b) Geometrical method for calculating the bone loss angle.}
    \label{mask_to_line_and_fig9}
\end{figure}

%% file: equations/eq4.tex
\begin{equation}
\theta = \cos^{-1}(u \cdot v)
\label{eq4}
\end{equation}

%% file: figures/fig20_TEX.tex
\begin{figure}[ht]
\centerline{\includegraphics[width=0.5\linewidth]{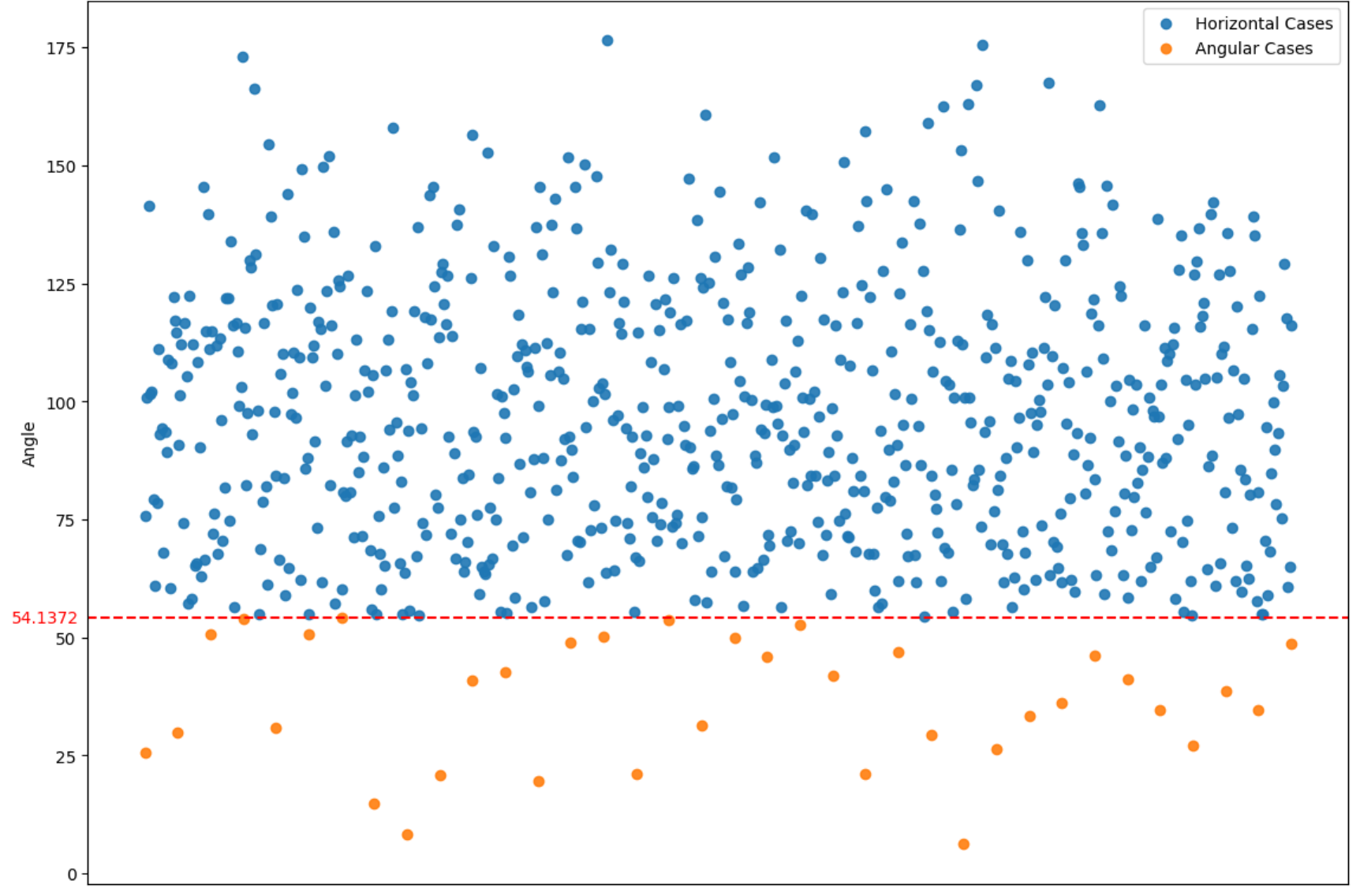}}
\captionsetup{justification=centering}
\caption{Angle values of ground-truth horizontal and angular cases.}
\label{fig20}
\end{figure}

%% file: tables/tooth_detection_results.tex
\begin{table}[ht]
    \centering
    \begin{tabular}{|c|c|c|c|c|}
        \hline
        \textbf{Model name} & \textbf{Data set type} & \textbf{Precision} & \textbf{mAP50} & \textbf{mAP50:95} \\
        \hline
        YOLOv8x & Train & 0.971 & 0.988 & 0.954 \\
        \cline{2-5}
        & Validation & 0.899 & 0.963  & 0.904 \\
        \cline{2-5}
        & Test & 0.892 & \textbf{0.963} & \textbf{0.907} \\
        \hline
        YOLOv8n & Train & 0.928 & 0.976 & 0.916 \\
        \cline{2-5}
        & Validation & 0.923 & 0.966 & 0.891 \\
        \cline{2-5}
        & Test & \textbf{0.918} & \textbf{0.963} & 0.894 \\
        \hline
        YOLOv9e & Train & 0.985 & 0.991 & 0.966 \\
        \cline{2-5}
        & Validation & 0.911 & 0.967 & 0.908 \\
        \cline{2-5}
        & Test & 0.913 & 0.960 & 0.904 \\
        \hline
    \end{tabular}
    \caption{\label{tooth_detection_results}Comparison of tooth detection results across different models. Evaluation metrics include precision, \gls{map50}, and \gls{map} across different \gls{iou} thresholds (from $0.5$ to $0.95$).}
\end{table}

%% file: figures/fig11_TEX.tex
\begin{figure}[!ht]
    \centering
    \includegraphics[width=0.5\linewidth]{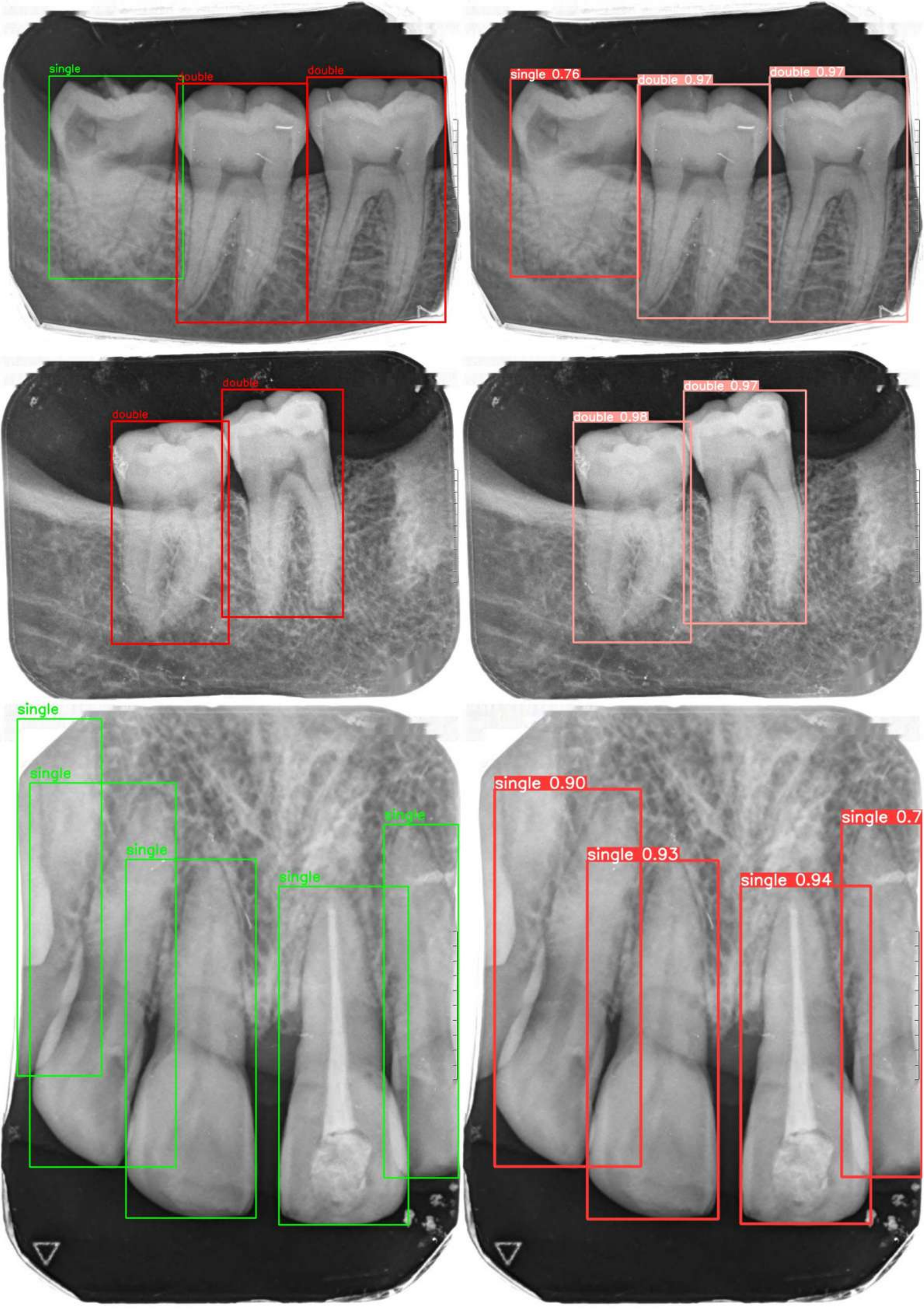}
    \captionsetup{justification=centering}
    \caption{Comparison of ground truth (left) and predicted (right) tooth detection.}
    \label{fig11}
\end{figure}

%% file: tables/keypoint_results.tex
\begin{table}[ht]
    \centering
    \resizebox{\textwidth}{!}{
    \begin{tabular}{|c|c|c|c|c|c|c|c|}
        \hline
        \textbf{Keypoint Type} & \textbf{Data set Type} &  \multicolumn{3}{|c|}{\textbf{Keypoint R-CNN}} & \multicolumn{3}{|c|}{\textbf{YOLOv8 Pose}} \\ 
        \cline{3-8}
        & & AP50:95 (OKS) & AP50 (OKS) & AP75 (OKS) & AP50:95 (OKS) & AP50 (OKS) & AP75 (OKS) \\
        \hline
        \multirow{3}{*}{CEJ} & Train & 0.958 & 0.958 & 0.958 & 0.849 & 0.851 & 0.849\\ 
        \cline{2-8}
         & Validation & 0.913 & 0.913 & 0.913 & 0.836 & 0.851 & 0.840\\ 
        \cline{2-8}
         & Test & \textbf{0.954} & \textbf{0.957} & \textbf{0.957} & 0.828 & 0.843 & 0.835\\ 
        \hline
        \multirow{3}{*}{\begin{tabular}[c]{@{}c@{}}Intersection point of \\ tooth boundary and alveolar \\ bone level\end{tabular}} & Train & 0.948 & 0.948 & 0.948 & 0.821 & 0.835 & 0.821\\ 
        \cline{2-8}
         & Validation & 0.913 & 0.913 & 0.913 & 0.818 & 0.844 & 0.815 \\ 
        \cline{2-8}
         & Test & \textbf{0.912} & \textbf{0.913} & \textbf{0.913} & 0.813 & 0.831 & 0.812\\ 
        \hline
        \multirow{3}{*}{Apex} & Train & 0.862 & 0.867 & 0.861 & 0.554 & 0.583 & 0.548\\ 
        \cline{2-8}
         & Validation & 0.847 & 0.855 & 0.848 & 0.470 & 0.505 & 0.466\\ 
        \cline{2-8}
         & Test & \textbf{0.815} & \textbf{0.821} & \textbf{0.815} & 0.498 & 0.547 & 0.478\\ 
        \hline
    \end{tabular}
    }
    \caption{Comparison of keypoint detection results across different models. The evaluation metric includes \gls{ap} across different \gls{oks} thresholds (from $0.5$ to $0.95$), \gls{ap} at \gls{oks} = $0.5$ (AP50 (OKS)), and \gls{ap} at \gls{oks} = $0.75$ (AP75 (OKS)).}
    \label{keypoint_results}
\end{table}

%% file: figures/fig12_and_fig13_and_fig14_TEX.tex
\begin{figure}[!ht]
    \centering
    \begin{subfigure}[t]{0.28\textwidth}
        \centering
        \includegraphics[width=\textwidth]{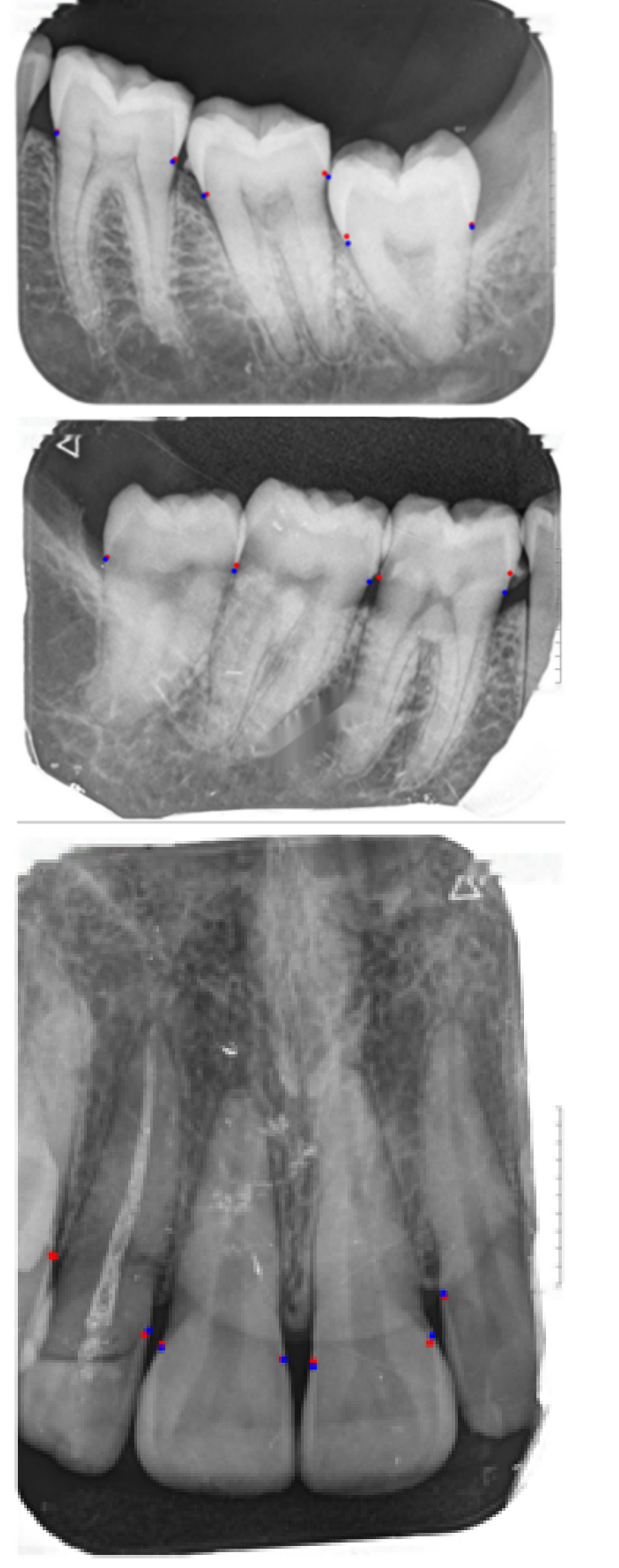}
        \caption{Cemento-enamel junction (CEJ) points.}
        \label{fig12_and_fig13_and_fig14:1}
    \end{subfigure}
    \hfill
    \begin{subfigure}[t]{0.28\textwidth}
        \centering
        \includegraphics[width=\textwidth]{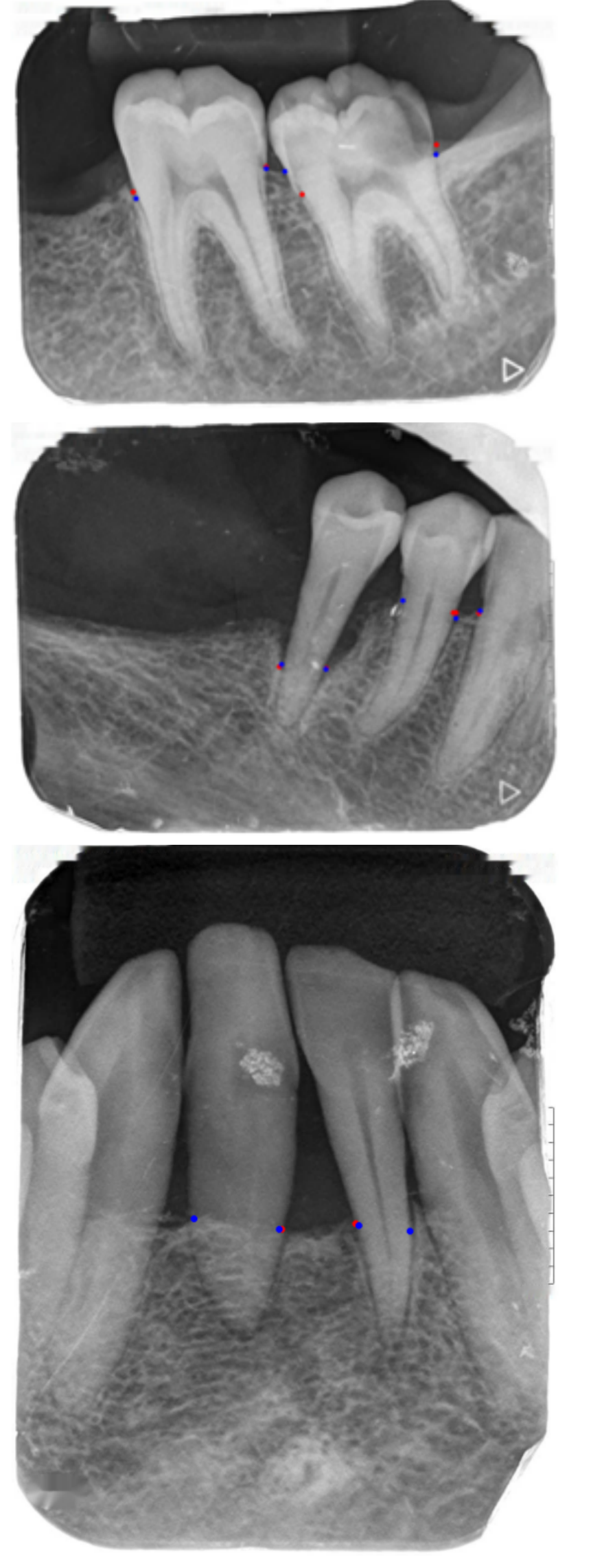}
        \caption{Alveolar bone level line and tooth intersection points.}
        \label{fig12_and_fig13_and_fig14:2}
    \end{subfigure}
    \hfill
    \begin{subfigure}[t]{0.28\textwidth}
        \centering
        \includegraphics[width=\textwidth]{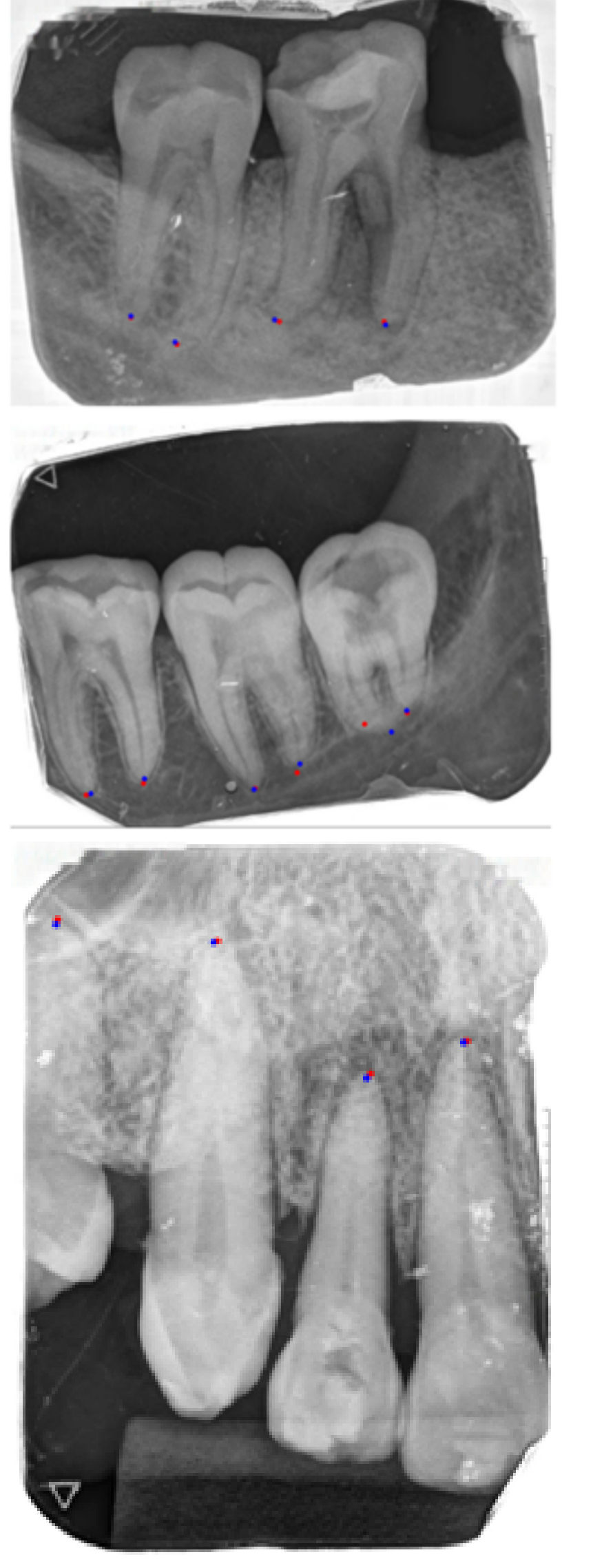}
        \caption{Apex points.}
        \label{fig12_and_fig13_and_fig14:3}
    \end{subfigure}
    \caption{Comparison of ground truth and predicted key points for tooth analysis: (a) cemento-enamel junction (CEJ) points; (b) alveolar bone level and tooth intersection points; (c) apex points. Blue points indicate ground truth, red points indicate predictions.}
    \label{fig12_and_fig13_and_fig14}
\end{figure}

%% file: figures/fig15_TEX.tex
\begin{figure}[ht]
    \centering
    \includegraphics[width=0.5\linewidth]{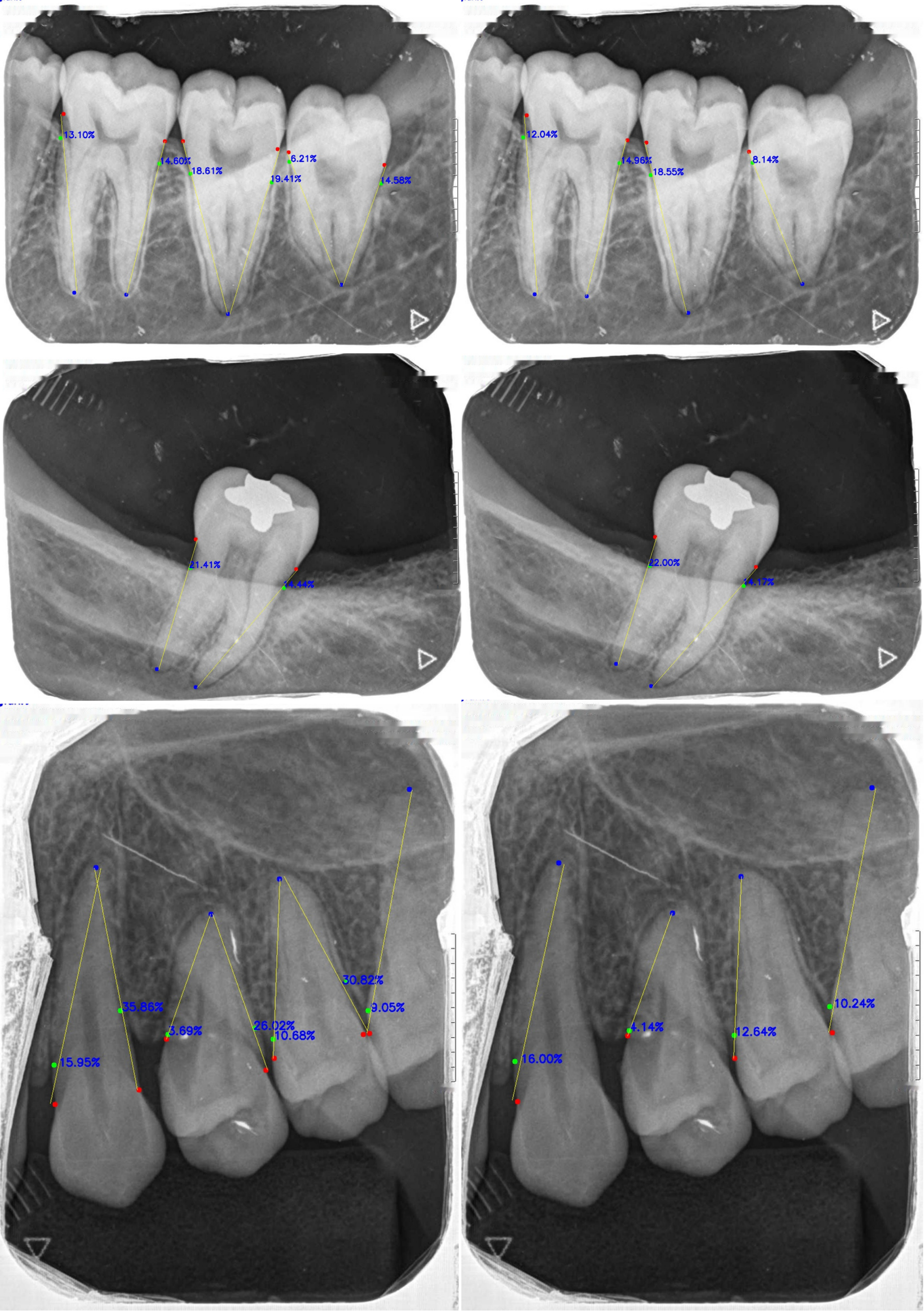}
    \caption{Comparison of ground truth (left) and predicted (right) alveolar bone loss severity. Red points represent the cementoenamel junction (CEJ), green points indicate the intersection between the alveolar crest bone level and the tooth surface, and blue points mark the apex of the tooth. Yellow lines represent the min-max line~\cite{cook_minmax_2023}. Bone loss severity are displayed on the corresponding sides of each tooth. Ground truth severity were calculated using manually annotated points, while predicted severity were derived from the model’s predicted points.}
    \label{fig15}
\end{figure}

%% file: tables/combined_mask_prediction_results.tex
\begin{table}[ht]
    \centering
    \begin{tabular}{|c|c|c|c|c|}
        \hline
        \textbf{Task} & \textbf{Model name} & \textbf{Data set type} & \textbf{AP50} & \textbf{AP50:95} \\
        \hline
        \multirow{6}{*}{Tooth Mask} 
        & \multirow{3}{*}{YOLOv8x-seg} & Train & 0.995 & 0.962 \\
        \cline{3-5}
        &  & Validation & 0.984 & 0.895 \\
        \cline{3-5}
        &  & Test & \textbf{0.978} & \textbf{0.900} \\
        \cline{2-5}
        & \multirow{3}{*}{Mask RCNN} & Train & 0.916 & 0.566 \\
        \cline{3-5}
        &  & Validation & 0.903 & 0.532 \\
        \cline{3-5}
        &  & Test & 0.885 & 0.548 \\
        \hline
        \multirow{6}{*}{Bone Level Mask} 
        & \multirow{3}{*}{YOLOv8x-seg} & Train & 0.697 & 0.202 \\
        \cline{3-5}
        &  & Validation & 0.555 & 0.144 \\
        \cline{3-5}
        &  & Test & \textbf{0.525} & \textbf{0.135} \\
        \cline{2-5}
        & \multirow{3}{*}{Mask RCNN} & Train & 0.033 & 0.009 \\
        \cline{3-5}
        &  & Validation & 0.011 & 0.004 \\
        \cline{3-5}
        &  & Test & 0.005 & 0.001 \\
        \hline
    \end{tabular}
    \caption{\label{combined_mask_prediction_results}Comparison of tooth and bone level mask prediction results across different models. Evaluation metrics include \gls{ap50} and \gls{ap} across \gls{iou} thresholds from $0.5$ to $0.95$.}
\end{table}

%% file: figures/fig16_and_fig17_TEX.tex
\begin{figure}[!ht]
    \centering
    \begin{subfigure}[t]{0.45\textwidth}
        \centering
        \includegraphics[width=\textwidth]{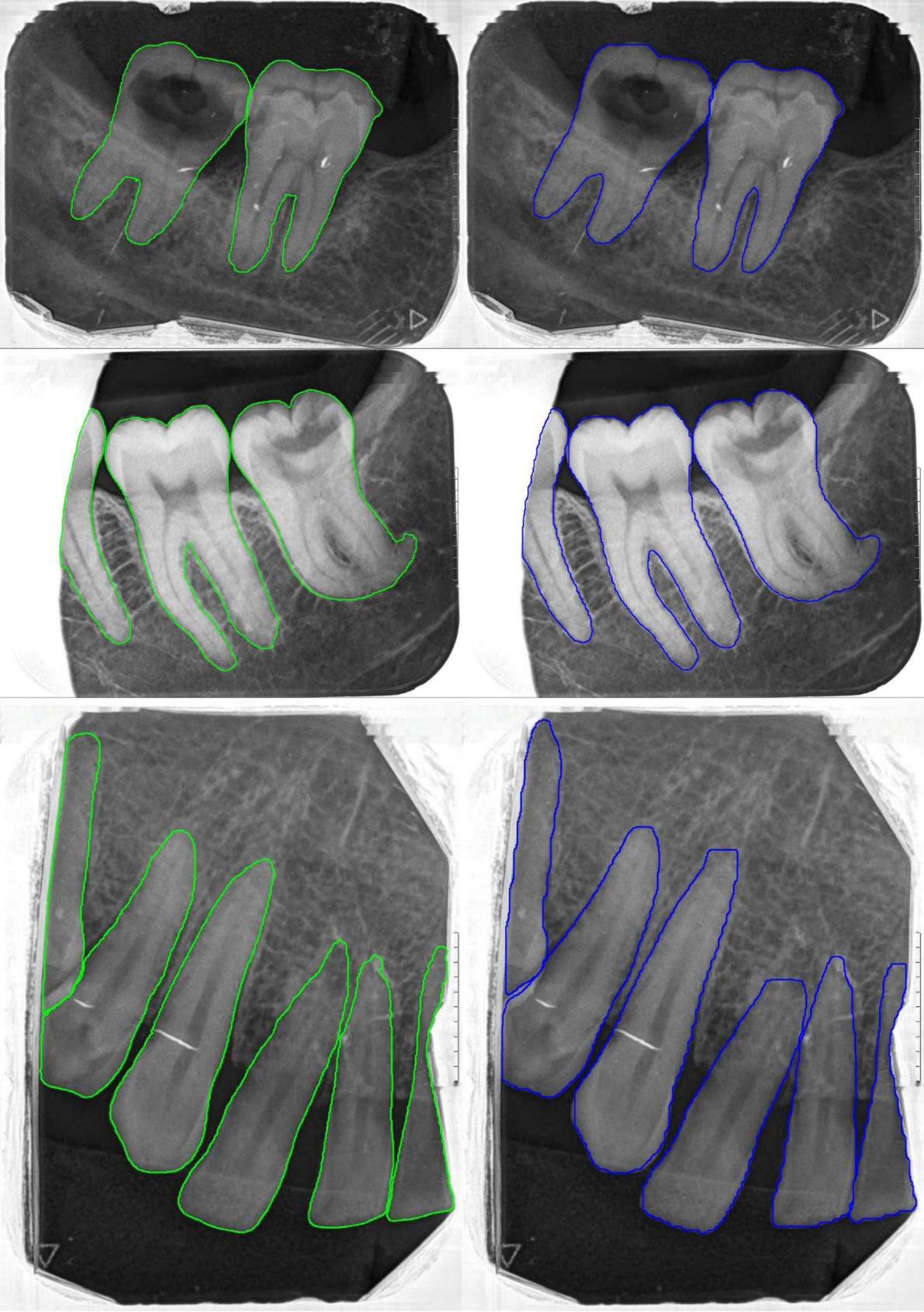}
        \caption{Comparison of ground truth (left) and predicted (right) tooth boundaries, taken from ground truth and predicted masks.}
        \label{fig16_and_fig17:1}
    \end{subfigure}
    \hfill
    \begin{subfigure}[t]{0.45\textwidth}
        \centering
        \includegraphics[width=\textwidth]{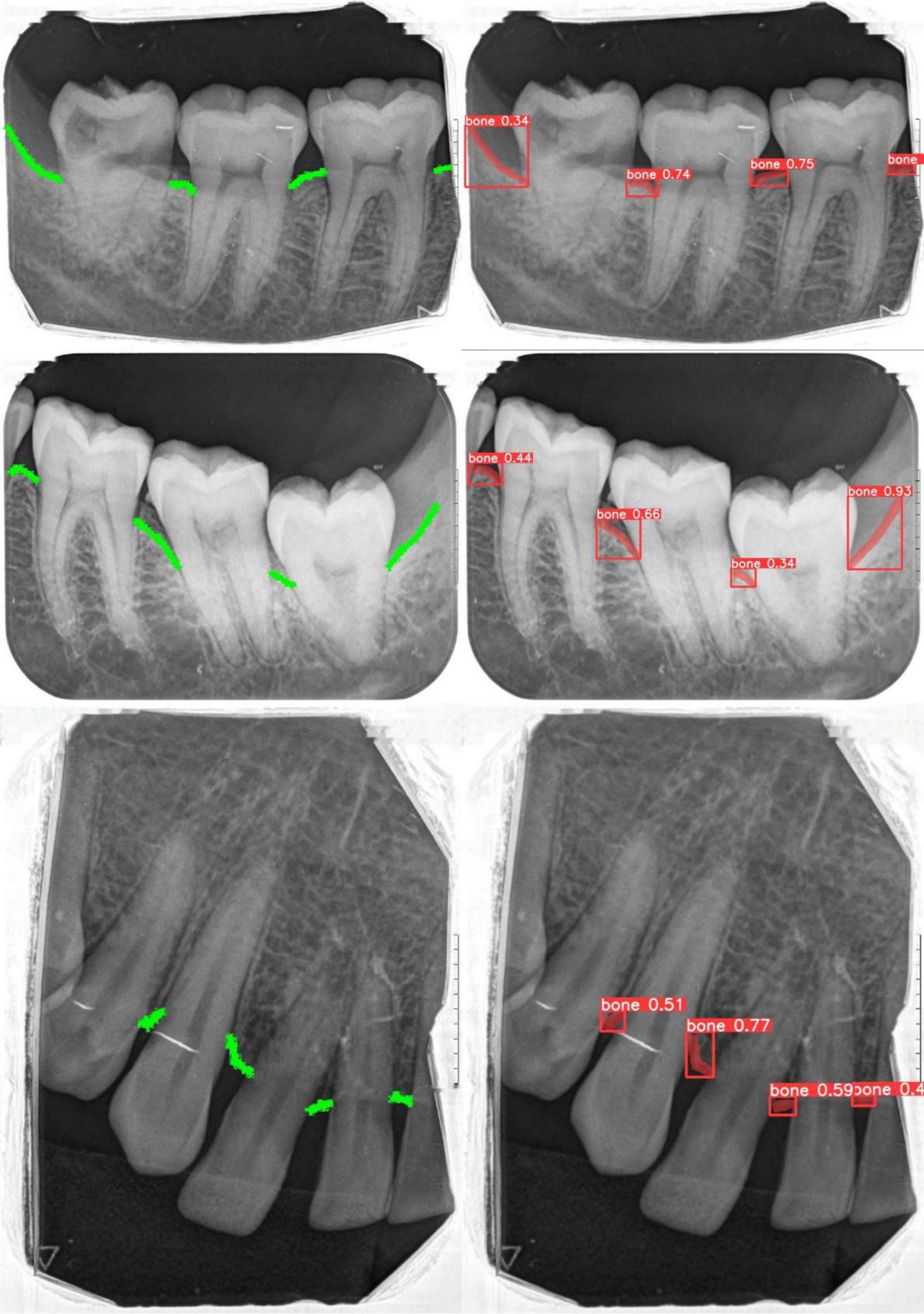}
        \caption{Comparison of ground truth (left) and predicted (right) bone line masks.}
        \label{fig16_and_fig17:2}
    \end{subfigure}
    \caption{Comparison of ground truth (left) and predicted (right) results: (a) tooth boundaries obtained from the corresponding masks; (b) bone line masks.}
    \label{fig16_and_fig17}
\end{figure}

%% file: tables/comparison_various_pixel_levels.tex
\begin{table}[ht]
    \centering
    \begin{tabular}{|c|c|c|c|}
        \hline
        \textbf{Thickness value} & \textbf{Train MSE} & \textbf{Validation MSE} & \textbf{Test MSE} \\ 
        \hline
        12px & 267.288 & 447.543 & 452.142 \\ 
        \hline
        11px & 291.570 & 510.671 & 456.658 \\ 
        \hline
        10px & 227.582 & 416.071 & \textbf{451.289} \\ 
        \hline
        9px & 340.573 & 569.294 & 612.147 \\ 
        \hline
        8px & 255.524 & 528.551 & 469.503 \\ 
        \hline
        7px & 556.906 & 555.431 & 923.583 \\ 
        \hline
        \end{tabular}
    \caption{\label{comparison_various_pixel_levels}Comparison of \gls{mse} between ground truth bone line and generated bone line across various pixel levels of masks.}   
\end{table}

%% file: figures/fig21_TEX.tex
\begin{figure}[!ht]
\centerline{\includegraphics[width=0.5\linewidth]{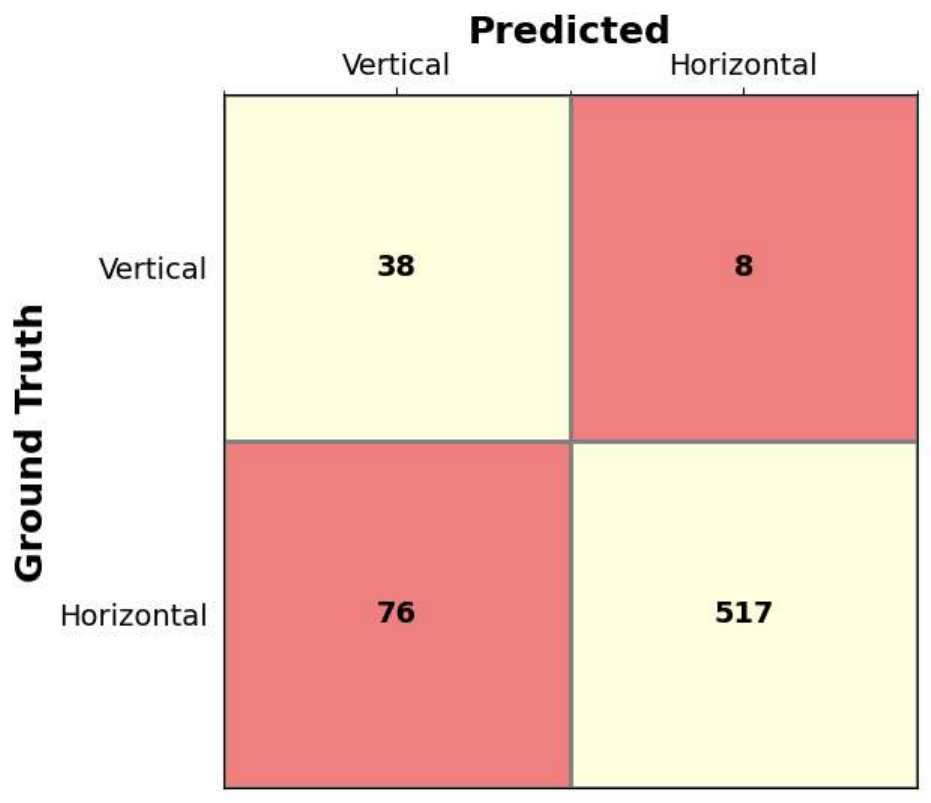}}
\captionsetup{justification=centering}
\caption{Confusion matrix of ground-truth and predicted horizontal and angular cases, based on $639$ total cases provided by dental professionals.}
\label{fig21}
\end{figure}

%% file: figures/fig18_TEX.tex
\begin{figure}[ht]
    \centering
    \includegraphics[width=0.5\linewidth]{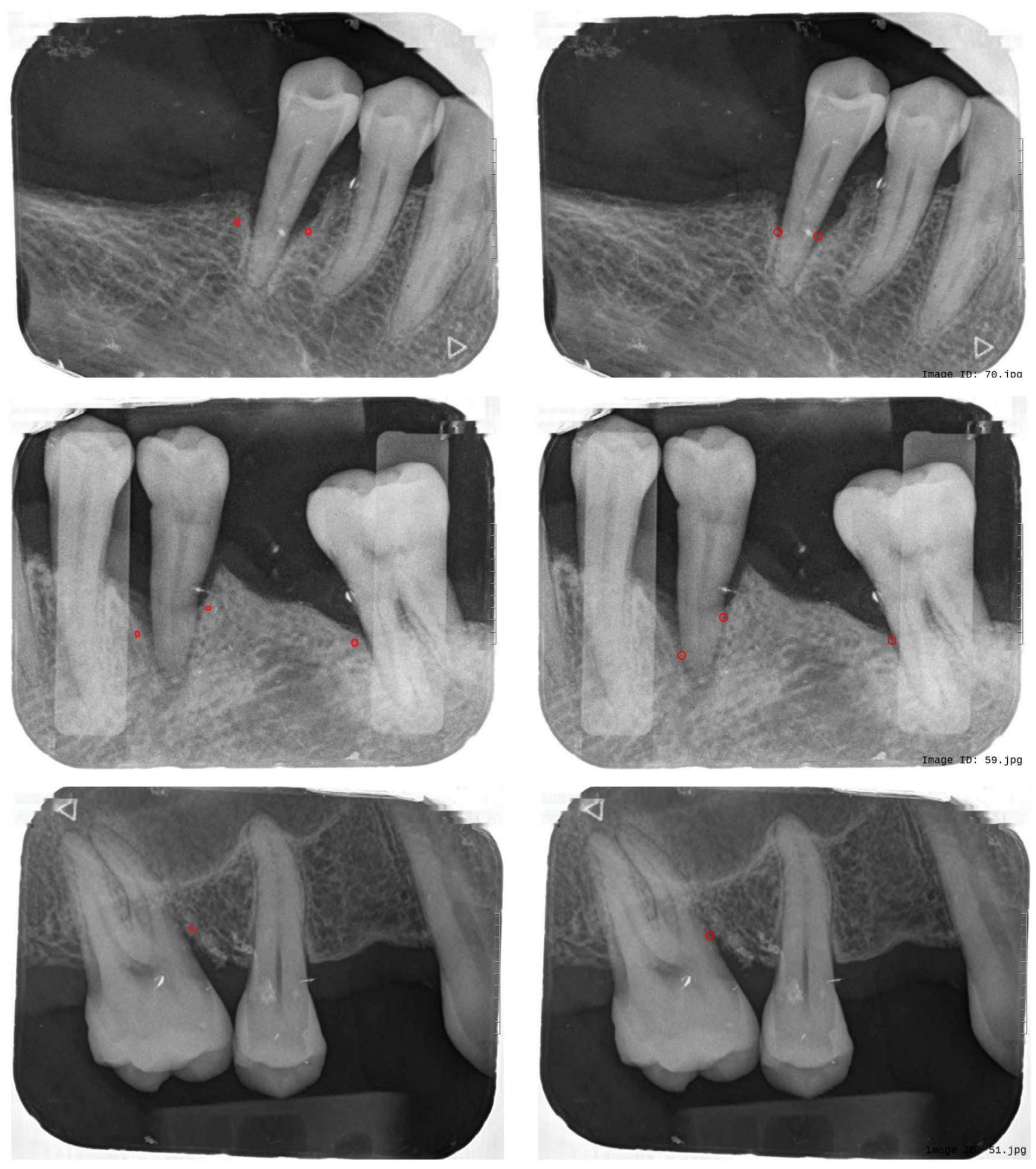}
    \captionsetup{justification=centering}
    \caption{Comparison of ground truth (left) and predicted (right) bone loss pattern. Red circles show the angular bone loss, while others show the horizontal bone loss cases.}
    \label{fig18}
\end{figure}